\documentclass{article}

\usepackage{PRIMEarxiv}

\usepackage[utf8]{inputenc} 
\usepackage[T1]{fontenc}    
\usepackage{hyperref}       
\usepackage{url}            
\usepackage{booktabs}       
\usepackage{amsfonts}       
\usepackage{nicefrac}       
\usepackage{microtype}      
\usepackage{lipsum}
\usepackage{fancyhdr}       
\usepackage{graphicx}       
\graphicspath{{figures/}}     

\usepackage[sort&compress]{natbib}

\usepackage{amssymb,amsmath,bm,amsfonts}
\usepackage{tabularray}
\usepackage{siunitx}
\usepackage{enumitem}
\usepackage{pifont}

\title{Physics-informed neural operator for parametric phase-field modelling of interfacial degradation and microstructural evolution
}

\author{
  Nanxi Chen$^{1}$,\;
  Rujin Ma$^{1,\ast}$,\;
  Airong Chen$^{1}$\\[8pt]
  $^{1}$College of Civil Engineering, Tongji University, Shanghai 200092, China\\
}

\begin{document}
\begingroup
\renewcommand{\thefootnote}{\fnsymbol{footnote}}
\footnotetext[1]{Correspondence to: \texttt{rjma@tongji.edu.cn}.}
\endgroup
\setcounter{footnote}{0}
\maketitle

\begin{abstract}
Predicting interfacial degradation and microstructural evolution through phase-field modelling is computationally intensive, particularly for parametric studies. While neural operators such as the Fourier neural operator (FNO) offer a promising route to acceleration, the absence of physical constraints compromises their generalisation and long-term stability. Here, we present PF-PINO, a physics-informed neural operator framework that embeds phase-field governing equation residuals directly into the training loss, together with gradient normalisation loss balancing and a staggered training scheme for coupled multi-field systems. PF-PINO is validated on four benchmarks including pencil-electrode corrosion, electro-polishing corrosion, dendritic solidification, and spinodal decomposition, collectively probing parametric generalisation across orders-of-magnitude kinetic regimes, non-periodic boundaries with stochastic morphologies, coupled multi-field dynamics with out-of-distribution extrapolation, and long-term autoregressive stability under complex pattern formation. Across all benchmarks, PF-PINO reduces relative $L^2$ errors by up to an order of magnitude over FNO and achieves sub-mesh-size interface accuracy, establishing a data-efficient surrogate for high-throughput parametric phase-field assessment in computational materials science and engineering applications.
\end{abstract}


\section{Introduction}
\label{sec:introduction}

The structural integrity and long-term performance of engineering systems are
critically influenced by interfacial degradation and microstructural evolution
phenomena that occur across multiple length and time scales. In service,
electrochemical corrosion and stress corrosion cracking progressively dissolve
metallic interfaces in bridges, offshore pipelines, and reinforced
infrastructure, representing one of the most economically and structurally
consequential degradation mechanisms in civil and mechanical engineering
\cite{maiPhaseFieldModel2016,cuiPhaseFieldFormulation2021,cuiGeneralisedMultiphasefieldTheory2022b}.
During processing, dendritic solidification governs the as-cast microstructure
of structural alloys, determining grain morphology, solute segregation, and
defect distribution that directly govern mechanical properties and failure
resistance
\cite{wei2016origin,quested2005grain,schmid2011thermodynamic,ditchek1979applications};
diffusion-driven phase separation
further controls the microstructural homogeneity and long-term phase stability
of alloys and composites and provides a route to spinodal architectures with
tunable mechanical properties
\cite{ramanarayanPhaseFieldStudy2003,xiangMorphologicalDesignTunable2024}.
Accurate quantitative prediction of these interfacial and microstructural
processes therefore constitutes a central challenge in computational mechanics,
underpinning materials design, manufacturing optimisation, and structural
integrity assessment. The phase-field method has emerged as a leading
computational framework for modelling such processes. Rooted in diffuse-interface
theory, it circumvents the need for explicit interface tracking by employing
continuous order parameters that evolve via thermodynamically consistent PDEs,
naturally capturing complex interfacial dynamics, topological changes, and
multi-physics couplings
\cite{steinbachPhasefieldModelsMaterials2009,boettingerPhasefieldSimulationSolidification2002,chenPhasefieldModelsMicrostructure2002}.
Despite its theoretical elegance and physical fidelity, the phase-field method
imposes prohibitive computational burdens in practice: high-fidelity simulations
require highly refined spatial discretisation and small time steps to resolve
sharp interface dynamics
\cite{zhangNumericalSolutionPhasefield2023,vedantam2006efficient,yangEfficientLinearStabilized2019},
often rendering the parametric studies and design optimisation workflows that
are essential in engineering practice computationally intractable.


In engineering practice, such parametric studies are often required for design evaluation and decision-making. The need to repeatedly simulate across broad parameter spaces thus creates a fundamental gap between the physical accuracy demanded by engineering applications and the computational tractability offered by conventional solvers. Machine learning offers a compelling avenue to bridge this gap through data-driven surrogate modelling \cite{liReviewApplicationsPhase2017a,wuSimGateDeepLearning2025,montesdeocazapiainAcceleratingPhasefieldbasedMicrostructure2021}. By training neural networks to approximate mappings from input parameters, such as material properties and initial or boundary conditions, to phase-field evolution trajectories, these methods enable rapid predictions that bypass the need for expensive numerical simulations. Various neural network architectures have been successfully deployed to capture the spatio-temporal dynamics of phase-field systems, including convolutional neural networks (CNNs) \cite{gaoCNNBasedSurrogatePhase2023,alhada-lahbabiMachineLearningSurrogate2023,peivasteMachinelearningbasedSurrogateModeling2022,alhada-lahbabiMachineLearningSurrogate2024}, recurrent neural networks (RNNs) \cite{huAcceleratingPhasefieldPredictions2022}, and long short-term memory (LSTM) networks \cite{montesdeocazapiainAcceleratingPhasefieldbasedMicrostructure2021,huAcceleratingPhasefieldSimulation2025,tiwariTimeSeriesForecasting2025}. Despite these advances, purely data-driven approaches remain fundamentally constrained by the quality and quantity of training data. Learning accurate solution mappings typically requires large-scale datasets comprising hundreds to thousands of high-fidelity simulation trajectories, incurring significant computational overhead that offsets the efficiency gains of surrogate modelling \cite{bonnevilleAcceleratingPhaseField2025,montesdeocazapiainAcceleratingPhasefieldbasedMicrostructure2021,huAcceleratingPhasefieldPredictions2022,alhada-lahbabiMachineLearningSurrogate2024}. When confronted with sparse or insufficiently diversified training data, these models often struggle to generalise to out-of-distribution parametric regimes \cite{liProbabilisticPhysicsinformedNeural2024,rabehBenchmarkingScientificMachinelearning2025,onetoInformedMachineLearning2025}. More critically, conventional data-driven training prioritises empirical data fidelity over physical consistency, often neglecting the underlying conservation laws and thermodynamic principles \cite{onetoInformedMachineLearning2025,karniadakisPhysicsinformedMachineLearning2021,valentePhysicsconsistentMachineLearning2025}, which can yield predictions that appear plausible yet violate essential physical constraints, leading to unreliable long-term evolution and unphysical artefacts that are particularly problematic in engineering applications.

Physics-informed machine learning offers a principled paradigm to address these shortcomings by embedding governing equations or other physical constraints directly into the learning process \cite{karniadakisPhysicsinformedMachineLearning2021,cuomoScientificMachineLearning2022}. A prominent example is physics-informed neural networks (PINNs) \cite{raissiPhysicsinformedNeuralNetworks2019}, which utilise automatic differentiation to evaluate partial differential equation (PDE) residuals at collocation points. By penalising these residuals as soft constraints within the loss function, PINNs can therefore be trained with sparse or even absent labelled data, relying on physical laws to regularise the learning process, and have been successfully applied to various phase-field problems relevant to materials degradation and microstructure evolution \cite{chenPFPINNsPhysicsinformedNeural2025b,chenSharpPINNsStaggeredHardconstrained2025,chen2025enforcinghiddenphysicsphysicsinformed,matteyNovelSequentialMethod2022,qiuPhysicsinformedNeuralNetworks2022,wight2020solving,manav2024phase}. However, PINNs typically represent solutions as coordinate-based neural networks that map spatio-temporal coordinates to field values for specific problem instances, requiring complete retraining for each new set of parameters \cite{kovachkiNeuralOperatorLearning2024}. This limitation renders PINNs unsuitable as general-purpose parametric solvers, severely limiting their utility in the parametric design and sensitivity studies that are central to engineering applications.

Neural operators address this limitation by learning mappings between infinite-dimensional function spaces, enabling generalisation across parameter regimes without retraining. Notable examples include the deep operator network (DeepONet) \cite{luLearningNonlinearOperators2021}, the Fourier neural operator (FNO), and its variants \cite{liFourierNeuralOperator2021,azizzadenesheliNeuralOperatorsAccelerating2024,kovachkiNeuralOperatorLearning2024}, all of which have demonstrated exceptional potential for capturing complex dynamics across varying parameter regimes in phase-field modelling \cite{oommenLearningTwophaseMicrostructure2022,ciesielskiDeepOperatorNetwork2025,bonnevilleAcceleratingPhaseField2025,xueEquivariantUShapedNeural2025}. While neural operators resolve the parametric generalisation limitation of PINNs, their standard data-driven training inherits the same fundamental shortcomings as conventional surrogates: heavy reliance on large labelled datasets and the absence of explicit physical constraints, which can compromise generalisation in out-of-distribution regimes and yield predictions that violate the underlying governing physics.

Physics-informed neural operators (PINOs) \cite{wangLearningSolutionOperator2021,liPhysicsInformedNeuralOperator2023} bridge this gap by embedding PDE residuals directly into the training of operator-based architectures, enabling solution operators that generalise across high-dimensional parametric spaces while maintaining physical consistency and substantially reducing reliance on exhaustive labelled datasets \cite{rosofskyApplicationsPhysicsInformed2023}. One representative example is the physics-informed DeepONet \cite{wangLearningSolutionOperator2021}, which penalises PDE residuals at randomly sampled collocation points via automatic differentiation. While successfully applied to 1D Allen--Cahn equations \cite{liTutorialsPhysicsinformedMachine2024} and gradient-flow-driven patterns \cite{liPhasefieldDeepONetPhysicsinformed2023}, physics-informed DeepONet faces inherent challenges in phase-field problems characterised by high-order spatial derivatives and sharp interface gradients: capturing intricate interface dynamics demands refined spatio-temporal sampling, which, coupled with the unfavourable scaling of automatic differentiation for higher-order derivatives \cite{luComprehensiveFairComparison2022}, incurs substantial memory overhead. Physics-informed FNO \cite{liPhysicsInformedNeuralOperator2023}, by contrast, leverages the fast Fourier transform for spatial convolutions and employs numerical or spectral differentiation to evaluate PDE residuals, offering a computationally efficient alternative well-suited to resolving fine interface details.

Despite this promise, the systematic application of physics-informed FNO to
phase-field modelling remains in its infancy. Recent work has explored
physics-informed FNO for specific coupled Allen--Cahn and Cahn--Hilliard
systems \cite{gangmeiLearningCoupledAllenCahn2025}, but these investigations
are confined to particular problem configurations and fall short of a
rigorous, cross-problem evaluation. Several critical questions remain
unanswered: whether physics-informed training consistently improves
generalisation across interpolation and out-of-distribution extrapolation
regimes; whether long-term autoregressive stability is reliably maintained;
and how Fourier-based architectures perform under the non-periodic boundary
conditions and sharp interface dynamics commonly encountered in engineering
phase-field problems. Furthermore, well-known computational challenges in
physics-informed training---namely gradient conflicts in coupled multi-field
systems \cite{wangGradientAlignmentPhysicsinformed2025a} and loss imbalance
between data fidelity and PDE residual terms
\cite{wangExpertsGuideTraining2023,chenGradNormGradientNormalization2018}---have
not been systematically addressed in this context.
These gaps motivate a systematic study of
physics-informed training for parametric phase-field modelling, together with
practical optimisation strategies for coupled multi-field systems.

In this work, we address these
gaps by developing PF-PINO, a physics-informed neural operator framework for
parametric phase-field modelling of interfacial degradation and
microstructural evolution. PF-PINO augments the data-driven training
objective with phase-field PDE residuals and incorporates optimisation
strategies for balancing and coordinating coupled residual losses. Using a
matched data-driven FNO as a controlled baseline, we investigate the effects of physics-informed training on predictive performance across representative phase-field settings. The main contributions are summarised as follows:
\begin{itemize}
    \item We establish representative parametric phase-field benchmarks
    covering broad parameter ranges, non-periodic boundaries, coupled
    anisotropic dynamics, out-of-distribution extrapolation, and stochastic
    coarsening.
    \item We incorporate gradient normalisation and staggered training into
    PF-PINO to balance competing loss terms and mitigate gradient conflicts in
    coupled multi-field systems.
    \item We demonstrate that PF-PINO consistently outperforms the matched FNO
    across all four benchmarks, reducing relative $L^2$ errors by up to an
    order of magnitude while improving interface accuracy, generalisation to
    unseen cases, long-term rollout stability, and physical consistency.
\end{itemize}

The remainder of this paper is organised as follows. Section~\ref{sec:methods} details the PF-PINO framework, including the neural operator architecture and the physics-informed training strategy. We then present the benchmark problems and evaluation results in Section~\ref{sec:results}, where we compare PF-PINO against standard FNO models and analyse the impact of physical constraints on model performance. Finally, we discuss the implications of our findings and potential future directions in Section~\ref{sec:conclusion}. The mathematical formulations, numerical implementation details, and hyperparameter settings for each benchmark are provided in the appendices for reproducibility and further reference.

\section{Methods}
\label{sec:methods}

\begin{figure*}[tbp]
    \centering
    \includegraphics[width=\textwidth]{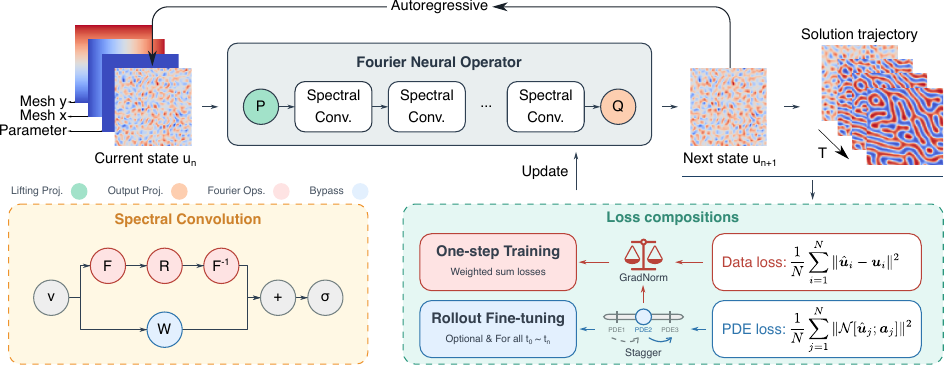}
    \caption{
    \textbf{Architecture and training strategy of the physics-informed neural operator framework for phase-field problems}.
    \textbf{Top}: The PF-PINO framework employs an autoregressive scheme to model temporal evolution of phase-field systems. The Fourier neural operator maps the current state $\bm{u}(\bm{x}, t_n)$ and static parameter fields $\bm{a}(\bm{x})$ to the subsequent state $\bm{u}(\bm{x}, t_{n+1})$ through lifting projection ($\mathcal{P}$), sequential spectral convolutions, and output projection ($\mathcal{Q}$). During inference, recursive application generates complete solution trajectories from initial conditions $\bm{u}(\bm{x}, t_0)$ to time $T$.
    \textbf{Bottom left}: Spectral convolution architecture. Input features undergo Fourier transformation ($\mathcal{F}$), multiplication with learnable spectral filters ($\mathcal{R}$) in frequency space, and inverse transformation ($\mathcal{F}^{-1}$). This spectral branch is combined with a linear bypass branch ($\mathcal{W}$) before nonlinear activation ($\sigma$), enabling efficient learning of solution operators in both spectral and physical domains.
    \textbf{Bottom right}: Composite loss function for physics-informed training. One-step training minimises weighted combination of data fidelity loss $\mathcal{L}_{\mathrm{d}}$ (measuring prediction accuracy against reference solutions) and PDE residual loss $\mathcal{L}_{\mathrm{p}}$ (enforcing governing equation compliance). Gradient normalisation and a staggered training scheme are employed to stabilise optimisation and mitigate gradient conflicts in coupled multi-field systems. Optional rollout fine-tuning further reduces accumulated PDE violations across the full temporal trajectory $t_0 \to t_n$.
    }
    \label{fig:framework}
\end{figure*}

Our PF-PINO framework integrates physics-informed constraints into the FNO architecture to learn parametric phase-field models. This framework employs an autoregressive scheme in which the neural operator maps the current system state $\bm{u}(\bm{x}, t_n)$ and static parameter fields $\bm{a}(\bm{x})$ to the subsequent state $\bm{u}(\bm{x}, t_{n+1})$, and complete solution trajectories are generated during inference by recursively applying this learned mapping from the initial condition. Unlike standard data-driven FNO training, PF-PINO explicitly embeds the residuals of the governing phase-field equations into the loss function \cite{liPhysicsInformedNeuralOperator2023}, where PDE residuals are computed via finite-difference or spectral differentiation and enforced as soft constraints, together with gradient normalisation loss balancing and a staggered training scheme for coupled multi-field systems, to ensure physical consistency alongside data fidelity. We provide an overview of the computational framework in Figure~\ref{fig:framework} and describe the architectural and training details in the subsequent sections. 


\subsection{Fourier neural operator architecture}

Neural operators enable learning of mappings between infinite-dimensional function spaces \cite{kovachkiNeuralOperatorLearning2024, azizzadenesheliNeuralOperatorsAccelerating2024}. Let $\mathcal{A}$ and $\mathcal{U}$ denote Banach spaces representing the input function space and the output solution space, respectively. These architectures approximate the solution operator $\mathcal{G}\colon \mathcal{A} \to \mathcal{U}$ through successive transformations: the input field $\bm{a}(\bm{x})$ is lifted to a high-dimensional representation, processed through $L$ layers combining local linear transformations $W_l$ with global integral kernel operators $\mathcal{K}_l$, and projected to the target solution space. This is mathematically expressed as:
\begin{align}
    \bm{v}_0(\bm{x}) &= \mathcal{P}(\bm{a}(\bm{x})), \\
    \bm{v}_{l+1}(\bm{x}) &= \sigma\left( \mathcal{W}_l \bm{v}_l(\bm{x}) + (\mathcal{K}_l \bm{v}_l)(\bm{x}) \right), \\
    \bm{u}(\bm{x}) &= \mathcal{Q}(\bm{v}_L(\bm{x})).
\end{align}
where $l=0,1,\ldots,L-1$ represents the layer index, $\mathcal{P}$ and $\mathcal{Q}$ are the input lifting and output projection operators, and $\sigma$ is a nonlinear activation function. The FNO \cite{liFourierNeuralOperator2021} is a specific instantiation of this architecture that implements the integral kernel operator $\mathcal{K}_l$ as a convolution in the Fourier domain, as shown in Figure~\ref{fig:framework} (bottom left). Exploiting the convolution theorem, the integral operation becomes a computationally efficient spectral convolution:
\begin{equation}
    (\mathcal{K}_l \bm{v}_l)(\bm{x}) = \mathcal{F}^{-1} \left( \mathcal{R}_l \cdot \mathcal{F}(\bm{v}_l) \right)(\bm{x}),
\end{equation}
where $\mathcal{F}$ and $\mathcal{F}^{-1}$ denote the Fourier transform and its inverse, respectively, and $\mathcal{R}_l$ represents learnable complex-valued weights that filter low-frequency Fourier modes.

\subsection{Physics-informed training strategy for parametric phase-field problems}

We consider the challenge of solving parametric, time-dependent phase-field problems governed by a general system of PDEs. Our objective is to learn a solution operator $\mathcal{G}: \mathcal{A} \to \mathcal{U}$ that maps input parameter fields $\bm{a} \in \mathcal{A}$ (comprising material properties and initial or boundary conditions) to the corresponding solution trajectories $\bm{u} \in \mathcal{U}$, such that:
\begin{equation}
    \mathcal{N}\left[\bm{u}(\bm{x},t); \bm{a}(\bm{x})\right] = 0, \quad \bm{x} \in \Omega, \quad t \in [0,T],
\end{equation}
where $\mathcal{N}$ denotes the differential operator representing the underlying physical laws.

To capture temporal evolution, we employ an autoregressive scheme. The temporal domain $[0,T]$ is discretised into sequential time steps $\{t_n\}_{n=0}^{N_t}$ with $t_0=0$ and $t_{N_t}=T$. The neural operator approximates the one-step temporal evolution mapping:
\begin{equation}
    \bm{u}(\bm{x}, t_{n+1}) = \mathcal{G}\left[\bm{u}(\bm{x}, t_n); \bm{a}(\bm{x})\right], \quad n=0,1,2,\ldots,
\end{equation}
where the operator $\mathcal{G}$ propagates the system state from time $t_n$ to $t_{n+1}$. During inference, the complete solution trajectory is generated by recursively applying the learned operator starting from the initial state $\bm{u}(\bm{x}, t_0)$.

The PF-PINO framework explicitly embeds the strong-form residuals of the governing equations into the training loss function. The composite loss function $\mathcal{L}$ is formulated as:
\begin{equation}
    \mathcal{L} = w_{\mathrm{d}} \mathcal{L}_{\mathrm{d}} + \sum_{q=1}^{N_{\text{pde}}} w_{\mathrm{p},q} \mathcal{L}_{\mathrm{p},q}, \label{eq:loss}
\end{equation}
where the individual components are given by:
\begin{align}
    \mathcal{L}_{\mathrm{d}} &= \frac{1}{N} \sum_{i=1}^{N} \| \hat{\bm{u}}_i - \bm{u}_i \|^2, \\
    \mathcal{L}_{\mathrm{p},q} &= \frac{1}{N} \sum_{j=1}^{N} \| \mathcal{N}_q[\hat{\bm{u}}_j; \bm{a}_j] \|^2.
\end{align}
Here, $N$ denotes the number of one-step input-output pairs, and each
$\bm{u}_i$ represents a complete solution field over the spatial grid. The
norms are evaluated over the complete spatial grid, with the data-loss norm
also including all variable channels.
$\mathcal{L}_{\mathrm{d}}$ minimises the discrepancy between one-step
predictions and reference solutions, while $\mathcal{L}_{\mathrm{p},q}$
penalises violations of the $q$-th governing equation. The derivative terms in
the PDE residuals are computed via finite-difference schemes or spectral
methods \cite{liPhysicsInformedNeuralOperator2023}, depending on the problem
context (see Appendix~\ref{sec:numerical-derivation-pino} for details).
$w_{\mathrm{d}}$ and $w_{\mathrm{p},q}$ are weighting coefficients that
balance the contributions of data fidelity and physical consistency during
training.

It is worth noting that the strong-form residual is well suited to the regular grids used here,
since spatial derivatives of the predicted fields can be evaluated directly without the test functions and numerical quadrature required by a weak-form formulation
\cite{kharazmiHpVPINNsVariationalPhysicsinformed2021,berroneVariationalPhysicsInformed2022}.
We use finite differences for the three non-periodic benchmarks and efficient spectral differentiation for the periodic spinodal decomposition benchmark.
Although coordinate-based automatic differentiation is well suited to PINNs with continuous coordinate inputs,
where collocation points can be evaluated in mini-batches
\cite{raissiPhysicsinformedNeuralNetworks2019,nabianEfficientTrainingPhysicsinformed2021},
applying it to spatial derivatives in FNO would require differentiating full-grid outputs with respect to full coordinate grids,
resulting in substantially greater computational and memory costs.

For applications demanding high precision, the framework incorporates an optional test-time fine-tuning stage. By minimising the PDE residuals over the predicted trajectory for specific test instances:
\begin{equation}
    \mathcal{L}_{\mathrm{p}} = \frac{1}{N_t}
    \sum_{m=0}^{N_t-1} \sum_{q=1}^{N_{\mathrm{pde}}}
    w_{\mathrm{p},q}
    \| \mathcal{N}_q\left[\hat{\bm{u}}(\bm{x}, t_m); \bm{a}(\bm{x})\right] \|^2,
\end{equation}
the model can be refined to reduce accumulated prediction errors and PDE violations over the rollout.

A theoretical interpretation of the PDE residual is provided in
Appendix~\ref{sec:residual-theory},
including its relation to one-step prediction error,
geometric regularisation of the learned solution graph,
and the propagation of local defects during autoregressive prediction.

\subsection{Gradient normalisation loss balancing}
The weighting coefficients $w_{\mathrm{d}}$ and $w_{\mathrm{p},q}$ that balance data fidelity and PDE residual terms can differ widely in magnitude and evolve at different rates during optimisation. To avoid domination by any single loss component, we adopt a gradient normalisation weighting strategy \cite{wangExpertsGuideTraining2023,chenGradNormGradientNormalization2018} that equalises their influence on parameter updates by matching gradient norms.

Concretely, let $\mathcal{J}$ index the loss components (data fidelity plus each PDE residual). At training step $s$ we compute the gradient norm of component $j$ with respect to model parameters $\bm\theta$ and form a scaling factor:
\begin{gather}
    g_j^{(s)} = \left\|\nabla_{\bm\theta} \mathcal{L}_j^{(s)}\right\|, \qquad j\in\mathcal{J}, \\
    \hat w_j^{(s)} = \frac{\sum_{k\in\mathcal{J}} g_k^{(s)}}{g_j^{(s)} + \epsilon},
\end{gather}
where $\|\cdot\|$ denotes the $L^2$ norm and $\epsilon$ is a small constant for numerical stability. The raw factors $\hat w_j^{(s)}$ are then smoothed using an exponential moving average to form the effective weights:
\begin{equation}
    w_j^{(s)} = \alpha_w\, w_j^{(s-1)} + (1-\alpha_w)\, \hat w_j^{(s)}, \qquad w_j^{(0)} = 1,
\end{equation}
where $\alpha_w\in[0,1)$ controls the smoothing. During training these weights scale each loss term so that components with smaller gradient norms are automatically amplified, promoting balanced optimisation between data fidelity and physical consistency.

\subsection{Staggered training scheme for coupled multi-field systems}

For coupled phase-field systems (e.g. Allen--Cahn/Cahn--Hilliard in corrosion or phase-field/heat diffusion in solidification), jointly enforcing all PDE residuals can produce competing gradient directions that slow or destabilise training. To mitigate these conflicts we adopt a staggered training scheme \cite{chenSharpPINNsStaggeredHardconstrained2025,wangGradientAlignmentPhysicsinformed2025a} in which the data fidelity loss remains active at every step, while PDE residuals for different fields are activated alternately.

Formally, for a system with $Q$ PDE residuals $\{\mathcal{L}_{\mathrm{p},q}\}_{q=1}^Q$ we introduce a one-hot gating vector $\delta^{(s)}$ with components $\delta_q^{(s)}\in\{0,1\}$ and $\sum_q\delta_q^{(s)}=1$. The total loss in Equation~\eqref{eq:loss} at iteration $s$ is reformulated as:
\begin{equation}
    \mathcal{L}^{(s)} = w_{\mathrm{d}}\,\mathcal{L}_{\mathrm{d}} + \sum_{q=1}^Q \delta_q^{(s)}\, w_{\mathrm{p},q}\,\mathcal{L}_{\mathrm{p},q}.
\end{equation}
The gating cycles through the $Q$ residuals with period controlled by the hyperparameter $N_{\text{switch}}$; one convenient implementation is
\begin{equation}
    \delta_q^{(s)} = \begin{cases}
        1, & q = 1 + \operatorname{mod}\big(\lfloor s / N_{\text{switch}} \rfloor,\,Q\big), \\
        0, & \text{otherwise},
    \end{cases}
\end{equation}
so that each residual is emphasised for $N_{\text{switch}}$ iterations before switching. Empirically this staggered protocol reduces gradient interference, improves stability, and accelerates convergence for strongly coupled multi-field problems \cite{chenSharpPINNsStaggeredHardconstrained2025}.

\section{Results}
\label{sec:results}
\begin{table*}[t]
    \centering
    \caption{Summary of benchmark results obtained using FNO and PF-PINO models. The relative $L^2$ errors are computed on the full spatio-temporal domain and averaged across all variable channels and test cases. The relative Hausdorff distance quantifies the accuracy of predicted interface locations for phase-field variables, computed at the final time step, normalised by the mesh size to provide a dimensionless measure of spatial accuracy, and averaged across all test cases.}
    \label{tab:benchmark_results}
    \begin{tblr}{
        width=\textwidth,
        colspec={Q[c,2.9]Q[c,2.8]Q[c,1]Q[c,1]Q[c,1]Q[c,1]},
        rowspec={Q[m]Q[m]Q[m]Q[m]Q[m]Q[m]},
        row{1}={font=\bfseries}}
        \hline
        \SetCell[r=2,c=1]{m} Benchmark test             &
        \SetCell[r=2,c=1]{m} Parametrisation            &
        \SetCell[r=1,c=2]{c} Rel. $L^2$ error (in $\%$) &                              &
        \SetCell[r=1,c=2]{c} Rel. Hausdorff distance    &                              &
        \\ \cline{3-4} \cline{5-6}
                                                        &                              & FNO   & PF-PINO & FNO  & PF-PINO \\
        \hline
        Pencil-electrode corrosion                      & Interface kinetics $L$       & 1.58  & 0.53    & 0.83    & 0.33 \\
        Electro-polishing corrosion                     & Initial interface profile    & 22.02 & 1.44     & 17.0   & 1.0 \\
        Dendritic crystal solidification                & Latent heat coeff. $K$       & 4.85  & 1.72     & 4.81  & 1.10  \\
        Spinodal decomposition                          & Mobility $M$ \& perturbation & 19.10 & 9.71    & 1.98  & 1.56   \\
        \hline
    \end{tblr}
\end{table*}
We evaluate the PF-PINO framework on four representative phase-field benchmarks: pencil-electrode corrosion, electro-polishing corrosion, dendritic crystal solidification, and spinodal decomposition. These benchmarks are deliberately selected to collectively probe a representative spectrum of computational challenges encountered in engineering phase-field modelling, with each case introducing progressively distinct demands on the surrogate model. The pencil-electrode corrosion benchmark provides a controlled 1D setting to assess parametric generalisation across corrosion kinetic regimes spanning several orders of magnitude. The electro-polishing benchmark extends this to 2D with non-periodic boundary conditions and stochastically parametrised initial morphologies, probing the robustness of Fourier-based architectures beyond their periodic assumption. The dendritic solidification benchmark introduces strongly coupled multi-field dynamics and anisotropic interface evolution, with test cases extending beyond the training parameter range to explicitly evaluate out-of-distribution extrapolation. Finally, the spinodal decomposition benchmark examines long-term autoregressive stability under stochastic initial conditions, where small perturbations evolve into complex coarsening microstructures that impose stringent demands on spectral fidelity. For each benchmark, high-fidelity reference datasets are generated using finite-element or spectral solvers, spanning broad parametric ranges to rigorously assess the model's accuracy, generalisation, and long-term predictive stability.

For a given parametric scenario, the phase-field PDEs are solved over a spatial domain discretised into $N_{\mathrm{grid}}$ grid points and a temporal horizon $[0,T]$ subdivided into $N_t$ time steps. Training data are constructed from one-step input-output pairs $(\bm{u}(\bm{x}, t_n), \bm{a}(\bm{x})) \to \bm{u}(\bm{x}, t_{n+1})$ for $n=0,1,\ldots,N_t-1$ across all $N_p$ trajectories, producing a total of $N_{\text{data}} = N_p \times N_t$ samples. These samples are randomly partitioned into training and validation subsets with a 75/25 split. Model performance is subsequently assessed on held-out test sets containing unseen parametric configurations. We evaluate the model's capacity to predict complete spatio-temporal solution trajectories through autoregressive rollouts from the initial condition $\bm{u}(\bm{x}, t_0)$ to the final time $T$.

We benchmark PF-PINO against a standard FNO baseline trained purely on data without the incorporation of physics-informed constraints.
The FNO is retained as a
matched backbone so that differences in accuracy, generalisation, and rollout
stability can be attributed to physics-informed training rather than changes
in architectural capacity. The PDE residual, gradient normalisation, and
staggered schedule operate at the loss and optimisation levels, independently
of the spectral-convolution layers, and can in principle be incorporated into
other grid-based neural operator backbones that predict spatial fields, such as
U-AFNO \cite{bonnevilleAcceleratingPhaseField2025} and E-UNO
\cite{xueEquivariantUShapedNeural2025}. Model accuracy is quantified using two primary metrics: the relative $L^2$ error and the relative Hausdorff distance
(see Appendix~\ref{sec:metrics} for definitions).
The
relative $L^2$ error measures the global discrepancy
between predicted and reference fields over the entire spatio-temporal domain,
providing an overall assessment of field accuracy.
To complement this global measure with interface-location accuracy,
which is central to interfacial degradation and microstructural evolution,
we compute the relative Hausdorff distance $d_\text{H}$
\cite{huttenlocherComparingImagesHausdorff1993,
tahaMetricsEvaluating3D2015}
between the predicted and reference interfaces at the final time step
and normalise it by the mesh size $\Delta x$
to obtain a dimensionless measure of spatial accuracy.
Both models are trained under identical hyperparameter settings and training protocols to ensure a fair comparison.
Table~\ref{tab:benchmark_results} presents a summary of these error metrics,
highlighting the comparative performance of PF-PINO and FNO.
The following sections provide detailed analyses of each benchmark case,
with supplementary information, including mathematical formulations, numerical implementation specifics,
physical parameters, and hyperparameter settings,
provided in the Appendices~\ref{sec:benchmark-description},
\ref{sec:numerical-implementation-dataset-generation},
\ref{sec:physical-parameters}, and \ref{sec:hyperparameters}.
Further supplementary results, including loss convergence curves,
ablation studies on training strategies, and additional visualisations of predicted trajectories,
are provided in Appendix~\ref{sec:additional-visualisations}.


\begin{figure*}[t]
    \centering
    \includegraphics[width=\textwidth]{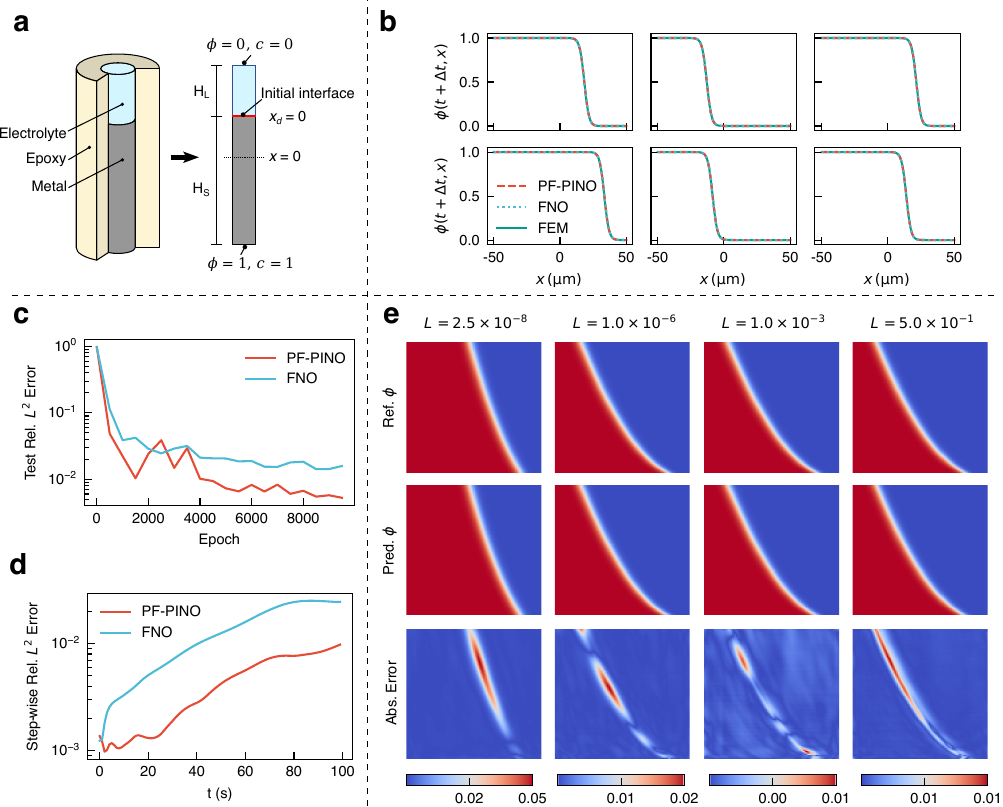}
    \caption{
        \textbf{Pencil-electrode corrosion simulation.}
        \textbf{a} Problem schematic of the 1D pencil-electrode corrosion model with parametric interface kinetics coefficient $L$.
        \textbf{b} Representative one-step predictions of phase-field variable $\phi$ (validation set) compared with FEM reference solutions.
        \textbf{c} Training convergence: relative $L^2$ error versus epoch for PF-PINO and FNO.
        \textbf{d} Autoregressive error accumulation: step-wise relative $L^2$ error evolution over the full prediction horizon.
        \textbf{e} Spatio-temporal phase-field distributions (reference, PF-PINO prediction, absolute error) for test cases with varying $L$ values. Spatial domain (horizontal axis) and temporal evolution (vertical axis) obtained through 100-step autoregressive rollouts.
    }
    \label{fig:corrosion1d}
\end{figure*}

\subsection{Pencil-electrode corrosion}

Corrosion-induced interfacial dissolution is among the most prevalent degradation mechanisms threatening the safety of metallic structures. Accelerated dissolution under aggressive electrochemical environments can dramatically shorten the service life of bridges, subsea pipelines, and port facilities, making parametric prediction of dissolution kinetics across varying electrochemical conditions essential for structural health monitoring and residual life assessment \cite{maiPhaseFieldModel2016,cuiGeneralisedMultiphasefieldTheory2022b}. This benchmark provides a controlled one-dimensional setting to assess \emph{parametric generalisation across corrosion kinetic regimes spanning several orders of magnitude}. We consider a pencil-electrode corrosion problem that models the evolution of a metal-electrolyte interface under electrochemical conditions (Figure~\ref{fig:corrosion1d}a), where a metal wire is immersed in an electrolyte solution and the corrosion process leads to dissolution and the formation of a moving interface. Following the Kim--Kim--Suzuki (KKS) phase-field formulation \cite{maiPhaseFieldModel2016,kimPhasefieldModelBinary1999,cuiPhaseFieldFormulation2021}, the metal-electrolyte interface is described by a continuous phase-field variable $\phi$, which smoothly transitions from 0 (electrolyte) to 1 (metal) and evolves according to the Allen--Cahn equation. The distribution of metal ions in the electrolyte is captured by a conserved, normalised concentration field $c$, governed by the Cahn--Hilliard equation. The coupled system is given by:
    \begin{align}
        \text{Allen--Cahn: } &   \frac{\partial \phi(\bm{x}, t)}{\partial t}=-L \frac{\delta \mathcal{E}}{\delta \phi},\label{eq:AC01} \\
        \text{Cahn--Hilliard: }    &
        \frac{\partial c (\bm{x}, t) }{\partial t}  =\nabla \cdot M \nabla \frac{\delta \mathcal{E}}{\delta c},\label{eq:CH01} 
    \end{align}
where $\mathcal{E}(\phi, c)$ is the total free energy functional of the system, $M$ is the mobility parameter controlling diffusion kinetics, and $L$ is the interface kinetics coefficient governing the rate of phase-field evolution. The specific form of the energy functional $\mathcal{E}$ and the associated variational derivatives are provided in Appendix~\ref{sec:benchmark-description}.

The initial phase-field profile is regularised as:
\begin{equation}
    \phi(x_d, 0) = \frac{1}{2}\left[
        1-\tanh\left(
        \sqrt{\frac{\omega_\phi}{2\alpha_\phi}} x_d
        \right)
        \right], \label{eq:initial_phi_profile}
\end{equation}
where $x_d$ denotes the distance from the initial interface, and $\omega_\phi$ and $\alpha_\phi$ are parameters determining the potential barrier and interfacial energy, respectively. The corresponding initial concentration profile is:
\begin{equation}
    c_0(x_d, 0) = h\left[\phi(x_d, 0)\right]c_{\mathrm{Se}}. \label{eq:initial_c_profile}
\end{equation}
Dirichlet boundary conditions are applied at the left boundary (metal) and right boundary (electrolyte), with $\phi=1, c=1$ at the left boundary and $\phi=0, c=0$ at the right boundary.

We parametrise the interface kinetics coefficient $L$ to enable predictions across a wide range of corrosion rates that directly affect the interface evolution. The dataset contains 11 scenarios with $L$ values sampled logarithmically from \num{1e-9} to \num{1.0}; reference solutions were computed using the finite-element method (FEM). To evaluate interpolation performance, five scenarios are used for training while the remaining six, unseen but drawn from the same range, form the test set. The central challenge lies in requiring the surrogate model to learn a single parametric mapping that accurately captures interface dissolution kinetics across corrosion rates spanning nine orders of magnitude in $L$, from activation-controlled to diffusion-controlled regimes, which produce vastly different spatio-temporal dynamics.

Figure~\ref{fig:corrosion1d}b compares one-step predictions from the validation set against FEM reference solutions, showing close agreement for both models. Training convergence profiles (Figure~\ref{fig:corrosion1d}c) reveal that PF-PINO achieves faster convergence and attains lower final errors, reflecting the regularising effect of the physics-informed constraints on the solution space. To evaluate long-term stability under autoregressive rollout, Figure~\ref{fig:corrosion1d}d tracks the evolution of relative $L^2$ errors across the entire temporal domain using models trained for \num{10000} epochs. Despite the inherent error accumulation characteristic of autoregressive schemes, PF-PINO consistently maintains substantially lower errors than FNO throughout all time steps. The spatio-temporal evolution and absolute error distributions across different $L$ values are visualised in Figure~\ref{fig:corrosion1d}e, demonstrating PF-PINO's capability to accurately capture interface dynamics across the full parametric space through 100-step autoregressive predictions.
To complement these field-based accuracy results,
thermodynamic dissipation during autoregressive prediction is further evaluated
through the normalised boundary-work-corrected energy change in
Figure~\ref{fig:corrosion1d_corrected_energy} in
Appendix~\ref{sec:pencil-electrode-corrosion-additional-visualisations}.

\subsection{Electro-polishing corrosion}

\begin{figure*}[tbp]
    \centering
    \includegraphics[width=\textwidth]{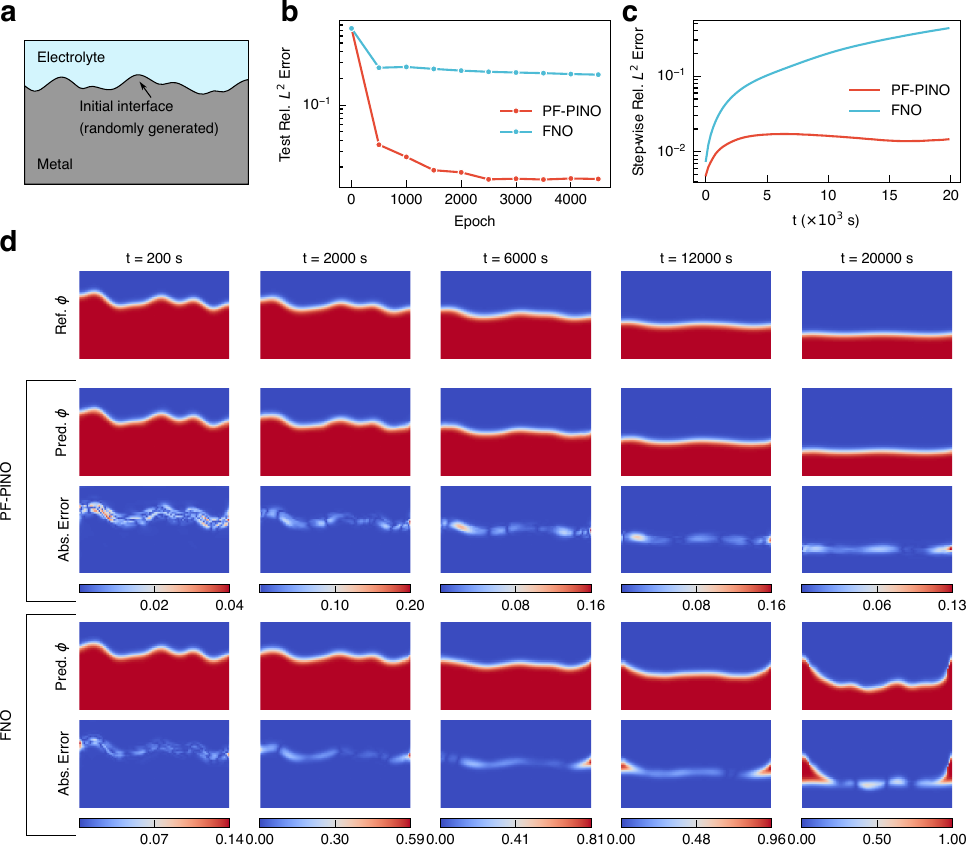}
    \caption{
        \textbf{Electro-polishing corrosion with parametric initial interface profiles.}
        \textbf{a} Problem schematic: 2D corrosion model with sinusoidally parametrised initial metal-electrolyte interface.
        \textbf{b} Training convergence: relative $L^2$ error versus epoch.
        \textbf{c} Autoregressive stability: step-wise error evolution showing PF-PINO's minimal accumulation versus FNO's divergence.
        \textbf{d} Spatial distributions of phase-field variable $\phi$ at final time for a representative test case (reference, PF-PINO and FNO predictions, absolute errors). Results obtained through 100-step autoregressive predictions highlight the accuracy and reduced boundary artefacts of PF-PINO compared to FNO.
    }
    \label{fig:corrosion2d}
\end{figure*}

Electro-polishing is a widely employed industrial technique for surface
finishing of metals via controlled electrochemical dissolution, with the
achieved surface quality directly governing the fatigue performance of
structural and aerospace components by controlling crack initiation sites under
cyclic loading \cite{landolt2003electrochemical}. Accurate prediction of the
evolving surface profile across diverse initial morphologies is therefore
important for optimising surface treatment processes and ensuring target
fatigue life in safety-critical applications. This benchmark extends to two
dimensions and introduces two compounding challenges: \emph{non-periodic
boundary conditions} that violate the periodicity assumption inherent in
Fourier-based architectures, and \emph{stochastically parametrised initial
morphologies} that probe generalisation across diverse interface geometries. We
consider the electro-polishing problem (Figure~\ref{fig:corrosion2d}a) following
the same KKS phase-field formulation as the pencil-electrode case
\cite{maiPhaseFieldModel2016}. To assess the model's ability to generalise
across diverse initial morphologies, we parametrise the metal--electrolyte
interface as a superposition of sinusoidal perturbations with varying
amplitudes and spatial frequencies.

Therefore, the initial interface profile is defined as:
\begin{equation}
    y_\text{int} = y_0 + \sum_{k=1}^{N_{\text{pert}}}a_k \cos\left(\pi k \xi\right),
\end{equation}
where $y_0$ is the mean interface position, $\xi\in[0,1]$ is the normalised coordinate along the interface, $N_{\text{pert}}$ is the number of modes, and $a_k$ denotes the amplitude of the $k$-th mode. The initial phase-field and concentration profiles are constructed from Eqs.~\eqref{eq:initial_phi_profile} and \eqref{eq:initial_c_profile} using the signed distance $y_d = y - y_{\text{int}}$. Boundary conditions are homogeneous Neumann on the lateral sides and Dirichlet on the vertical boundaries (top electrolyte: $\phi=0,\; c=0$; bottom metal: $\phi=1,\; c=1$). These non-periodic boundary conditions break the spatial periodicity assumed by Fourier-based architectures and can introduce boundary artefacts that degrade FNO performance \cite{liFourierNeuralOperator2021,kovachkiNeuralOperatorLearning2024}. We mitigate edge effects using padding techniques, but residual artefacts can still arise near the boundaries, making this benchmark a stringent test of model robustness to non-periodic boundaries. We generate 15 interface profiles with randomly sampled parameters, using 10 for training and 5 unseen configurations for testing.

Figure~\ref{fig:corrosion2d}b demonstrates that PF-PINO rapidly achieves relative $L^2$ errors over an order of magnitude lower than FNO during training. Autoregressive rollout performance (Figure~\ref{fig:corrosion2d}c) reveals that PF-PINO maintains stable convergence with minimal error accumulation, whereas FNO exhibits substantial error growth over time. Representative predictions for a test case are shown in Figure~\ref{fig:corrosion2d}d, where PF-PINO accurately reproduces the corrosion dynamics across the entire domain. Notably, although padding techniques are employed to mitigate boundary artefacts, FNO still exhibits significant errors near the non-periodic boundaries.
These results demonstrate that physics-informed training not only improves generalisation across parametric initial morphologies but also mitigates the boundary inconsistencies inherent to Fourier-based architectures, which is particularly relevant for realistic corrosion geometries that rarely conform to periodic boundary conditions.
To quantify
computational cost, we provide a wall-clock comparison of PF-PINO and FNO
training, PF-PINO inference, and FEM simulation in
Figure~\ref{fig:corrosion2d_time_cost} in
Appendix~\ref{sec:electro-polishing-corrosion-additional-visualisations}.
To further assess the effect of the autoregressive time step,
we provide an additional sensitivity
analysis for $\Delta t=\qtyrange{50}{400}{\second}$ in
Figure~\ref{fig:corrosion2d_dt_sensitivity} in
Appendix~\ref{sec:electro-polishing-corrosion-additional-visualisations}.
For more trajectory visualisations in the test set, see Figure~\ref{fig:corrosion2d_test_sol_samples} in Appendix~\ref{sec:electro-polishing-corrosion-additional-visualisations}.



\subsection{Dendritic crystal solidification}
\begin{figure*}
    \centering
    \includegraphics[width=\textwidth]{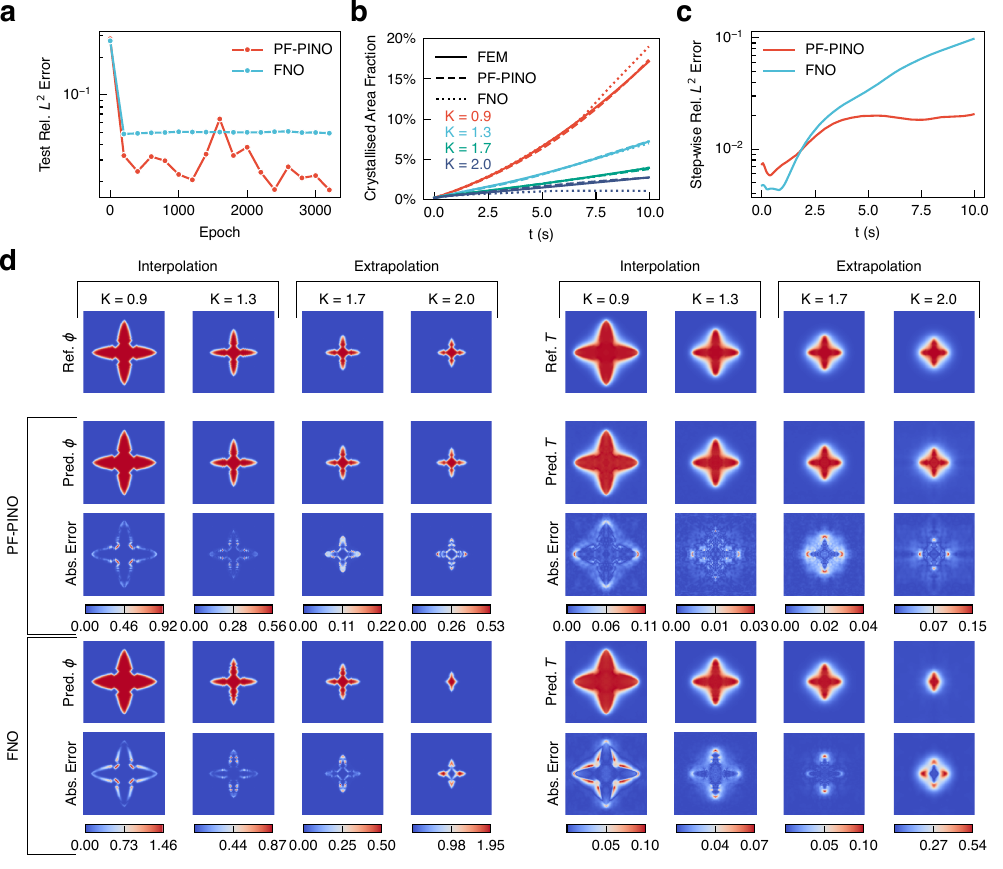}
    \caption{
        \textbf{Dendritic crystal solidification with parametrised latent heat coefficient.}
        \textbf{a} Training convergence: averaged relative $L^2$ error for phase-field variable $\phi$ and temperature $T$.
        \textbf{b} Crystallised area fraction evolution for different latent heat coefficients $K$, comparing PF-PINO and FNO predictions with FEM reference.
        \textbf{c} Autoregressive error accumulation: step-wise relative $L^2$ error showing PF-PINO's stability versus FNO's divergence.
        \textbf{d} Spatial distributions of $\phi$ and $T$ at final time ($t=\qty{10.0}{s}$) for test cases (reference, predictions, absolute errors). Results for $K=0.9,\,1.3$ (interpolation) and $K=1.7,\,2.0$ (extrapolation) demonstrate PF-PINO's superior accuracy in both regimes.
    }
    \label{fig:solidification}
\end{figure*}

The morphology and scale of dendritic microstructures formed during solidification are key determinants of the mechanical properties of cast structural alloys---finer dendrite arm spacings and reduced solute segregation yield superior strength, toughness, and fatigue resistance, while coarser microstructures promote premature failure initiation \cite{wei2016origin,quested2005grain,schmid2011thermodynamic}. Reliable prediction of solidification microstructure as a function of processing conditions thus directly supports casting process optimisation and alloy design for load-bearing structural applications. This benchmark targets \emph{strongly coupled multi-field dynamics with anisotropic interface evolution} and, crucially, includes test cases that extend beyond the training parameter range to explicitly evaluate \emph{out-of-distribution extrapolation}. We examine dendritic crystal growth, a ubiquitous phase transformation process that involves the formation of complex branched structures driven by anisotropic interfacial energy. The governing equations couple the Allen--Cahn equation for phase-field evolution with heat diffusion. The phase-field variable $\phi(\bm{x}, t)$ distinguishes solid ($\phi=1$) from liquid ($\phi=-1$) phases, while the dimensionless temperature field $T(\bm{x}, t)$ characterises the thermal distribution. The governing equations are \cite{karmaPhasefieldModelDendritic1999,yangEfficientLinearStabilized2019}:
    \begin{gather}
        \rho(\phi) \frac{\partial \phi}{\partial t} = -\frac{\delta \mathcal{E}}{\delta \phi} - \frac{\lambda}{\varepsilon} h'(\phi) T, \label{eq:solid_ac} \\
        \frac{\partial T}{\partial t} = \nabla \cdot (D \nabla T) + K h'(\phi) \frac{\partial \phi}{\partial t}, \label{eq:solid_heat}
    \end{gather}
where $\rho(\phi)$ denotes the phase-dependent mobility, $\varepsilon$ controls interface width, $\lambda$ represents the kinetic coefficient, $D$ is the thermal diffusivity, and $K$ is the latent heat parameter governing solidification kinetics. The energy functional $\mathcal{E}(\phi, T)$ incorporates anisotropic gradient energy with directional dependence parametrised by $\kappa(\nabla\phi) = 1 + \sigma\cos(m\theta)$, where $m$ determines the crystallographic symmetry and $\sigma$ controls anisotropy strength. The function $h(\phi) = \frac{1}{5}\phi^5 - \frac{2}{3}\phi^3 + \phi$ represents latent heat generation at the interface. Equation \eqref{eq:solid_ac} governs interface evolution driven by variational minimisation of anisotropic free energy and undercooling, while Equation \eqref{eq:solid_heat} describes heat diffusion with source terms from latent heat release. The anisotropic gradient terms in $\delta E/\delta\phi$ enable formation of dendritic morphologies characteristic of crystallisation. Complete expressions for the variational derivatives are provided in Appendix~\ref{sec:benchmark-description}.

We initialise the system with a circular solid seed of radius $r_0=0.05$ at the centre of a supercooled liquid melt. The initial phase-field and temperature profiles are given by:
    \begin{gather}
        \phi(x,y,0) = \tanh\left(\frac{r_0 - \sqrt{x^2 + y^2}}{\sqrt{2}\varepsilon}\right), \\
        T(x,y,0) = \begin{cases}
            0,    & \sqrt{x^2 + y^2} < r_0, \\
            -0.6, & \sqrt{x^2 + y^2} \geq r_0.
        \end{cases}
    \end{gather}
Homogeneous Neumann boundary conditions are applied for both $\phi$ and $T$ on all domain boundaries. To evaluate model performance across different solidification rates, we parametrise the dimensionless latent heat coefficient $K$, which controls the rate of heat release during phase transformation. The training set contains five $K$ values uniformly sampled in $[0.8,\,1.6]$. The test set includes four unseen values chosen to explicitly evaluate both interpolation and extrapolation: $K=0.9$ and $1.3$ (interpolation within the training range) and $K=1.7$ and $2.0$ (extrapolation beyond the training range). This split allows a direct assessment of the model's robustness to out-of-distribution parameter values.

Training convergence profiles (Figure~\ref{fig:solidification}a) reveal that FNO exhibits premature stagnation in error reduction for both phase-field variable $\phi$ and temperature $T$, whereas PF-PINO achieves substantially lower final errors through sustained convergence. To evaluate the model's capacity to capture global solidification kinetics, Figure~\ref{fig:solidification}b tracks the evolution of crystallised area fraction throughout the prediction horizon. PF-PINO faithfully reproduces the FEM reference trajectory across all $K$ values and time steps, while FNO displays pronounced deviations, particularly during the later stages of solidification. The autoregressive error evolution (Figure~\ref{fig:solidification}c) further confirms that PF-PINO maintains stable errors with minimal accumulation, while FNO errors progressively diverge over extended rollouts. Spatial predictions at the final time step (Figure~\ref{fig:solidification}d) demonstrate that PF-PINO accurately captures the intricate dendritic morphology for both interpolated ($K=0.9,\,1.3$) and extrapolated ($K=1.7,\,2.0$) parameter values. In stark contrast, FNO exhibits substantial discrepancies at higher $K$ values, particularly in the extrapolation regime, underscoring the critical role of physics-informed constraints in regularising the solution space and enabling robust generalisation beyond the training distribution. A complete $2\times2$ ablation of gradient normalisation and staggered training is reported in Figure~\ref{fig:solidification_test_l2_ablation} and Table~\ref{tab:solidification_ablation}, confirming that both strategies improve convergence and long-term rollout accuracy. Additional trajectory visualisations and experiments on the data-efficiency of PF-PINO are also provided in Appendix~\ref{sec:dendritic-solidification-additional-visualisations}

\subsection{Spinodal decomposition}

Spinodal decomposition can significantly affect the long-term microstructural stability of certain high-strength structural and nuclear alloys; progressive phase separation from a nominally homogeneous state generates compositional fluctuations and microstructural heterogeneities that commonly reduce ductility and impact toughness, alter strength and hardness, and may ultimately promote embrittlement failure \cite{ramanarayanPhaseFieldStudy2003,krekhovPhaseSeparationPresence2004}. Accurate prediction of decomposition kinetics across alloy compositions is therefore critical for assessing long-term mechanical reliability and establishing safe service limits under elevated-temperature or irradiation environments. The final benchmark examines \emph{long-term autoregressive stability and spectral fidelity} under stochastic initial conditions, where small perturbations evolve into complex coarsening microstructures that impose the most stringent demands on the surrogate model. We study spinodal decomposition in binary alloys, which models spontaneous phase separation from an initially homogeneous state with small random perturbations, evolving towards distinct equilibrium phases through diffusion-driven coarsening. This process is governed by the Cahn--Hilliard equation, a conserved-order-parameter formulation that describes the temporal evolution of the normalised concentration field $c(\bm{x}, t)$ and its conjugate chemical potential field $\mu(\bm{x}, t)$. The governing equations are \cite{krekhovPhaseSeparationPresence2004,ramanarayanPhaseFieldStudy2003}:
    \begin{gather}
        \frac{\partial c}{\partial t} = \nabla \cdot \left( M \nabla \mu \right), \label{eq:ch_conservation} \\
        \mu = f'(c) - \lambda \Delta c, \label{eq:ch_potential}
    \end{gather}
where $M$ denotes the mobility parameter controlling diffusion kinetics, $\lambda = \epsilon^2$ represents the gradient energy coefficient with interface width parameter $\epsilon$, and $f'(c) = c^3 - c$ is the derivative of the double-well free energy density.

\begin{figure*}[tbp]
    \centering
    \includegraphics[width=\textwidth]{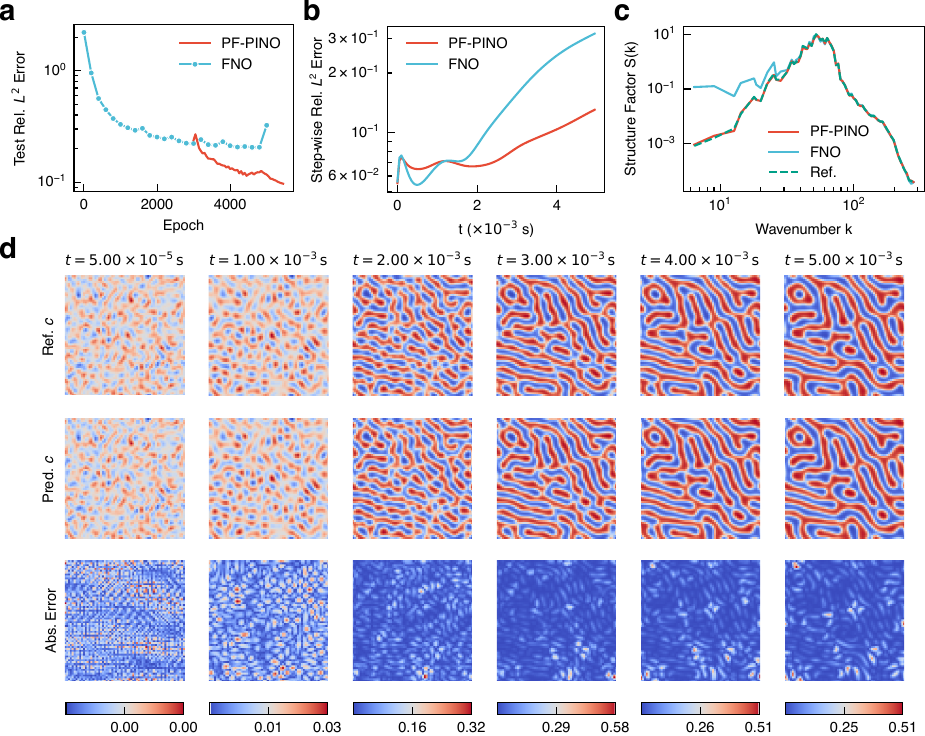}
    \caption{
        \textbf{Spinodal decomposition with parametrised mobility and initial perturbations.}
        \textbf{a} Training convergence: relative $L^2$ error versus epoch, illustrating data-driven training followed by physics-informed fine-tuning.
        \textbf{b} Autoregressive error evolution: step-wise relative $L^2$ error demonstrating the superior stability of PF-PINO.
        \textbf{c} Radially averaged structure factor $S(k)$ at the final time step averaged across all test cases. The close alignment of the PF-PINO profile with the reference in the low-$k$ regime highlights its superior capture of global thermodynamic features.
        \textbf{d} Spatial distributions of $c$ at sequential time steps for the test case with $M=1.2$ (reference, predictions, absolute errors). The results show the accurate capture of phase separation and domain coarsening dynamics by PF-PINO.
    }
    \label{fig:spinodal}
\end{figure*}

The central challenge in this benchmark lies in the model's ability to maintain predictive accuracy over long autoregressive rollouts while faithfully reproducing the spectral characteristics of the evolving microstructure. Small initial perturbations are amplified and reorganised through nonlinear dynamics into complex, coarsening microstructures, where even minor spectral errors can accumulate and cascade across scales, imposing stringent requirements on both long-horizon stability and spectral fidelity. To assess robustness across kinetic regimes and initial conditions, we parametrise both the mobility coefficient $M$ and the spectral content of the initial perturbations superimposed on the homogeneous concentration field. The mobility $M$ is sampled uniformly from $[0.5,\,1.5]$, and the initial perturbations are generated as a band-limited random field defined by:
\begin{equation}
    c(x,y,0) = c_0 + \delta c(x,y),
\end{equation}
where $\delta c(x,y)$ represents random perturbations of small amplitude, generated using a band-limited Fourier series:
\begin{equation}
    \delta c(x,y) = \sum_{i=1}^{N_{\text{pert}}}
    a_i \cos\left(2\pi (k_{x,i} x + k_{y,i} y) + \phi_i\right).
\end{equation}
Here, $N_{\text{pert}}$ denotes the number of perturbation modes, while $a_i$ and $\phi_i$ are randomly sampled amplitudes and phases, and $k_{\text{min}} \leq k_{x,i}, k_{y,i} \leq k_{\text{max}}$ are the bounded wave numbers that control the spatial frequency of the perturbations. The mean concentration is set to $c_0=0$ to induce a symmetric phase separation scenario.

The dataset comprises 25 scenarios, with 20 allocated for training and validation and 5 reserved for testing. For this benchmark, we employ a two-stage training strategy: the model is first trained on one-step input-output pairs from reference data, followed by physics-informed fine-tuning over the complete autoregressive rollout trajectory to further minimise accumulated PDE residuals across all time steps.

Training convergence (Figure~\ref{fig:spinodal}a) demonstrates that the initial data-driven phase achieves rapid error reduction for both FNO and PF-PINO. Subsequent physics-informed fine-tuning yields substantial additional improvement for PF-PINO. Autoregressive rollout performance (Figure~\ref{fig:spinodal}b) shows that PF-PINO maintains consistently lower errors throughout the prediction horizon, whereas FNO errors remain significantly higher. To further quantify the model's ability to capture statistical microstructural features, we compute the radially averaged structure factor $S(k)$ at the final time step \cite{cook1970brownian}, averaged across all test cases (Figure~\ref{fig:spinodal}c). Notably, PF-PINO exhibits significantly higher fidelity across the full spectral range compared to FNO, which exhibits deviations from the reference spectrum, particularly in the low-wave-number regime. This discrepancy reflects the inherent limitation of FNO's truncated Fourier representation, where spectral errors can accumulate and cascade across scales during autoregressive rollouts. The physics-informed loss mitigates this effect by enforcing PDE residuals that implicitly anchor predictions across the full spectrum. Representative spatial predictions for a test case with $M=1.2$ (Figure~\ref{fig:spinodal}d) reveal that PF-PINO accurately captures the characteristic phase separation dynamics, including domain nucleation, growth, and coarsening across multiple time steps.
Additional final-time predictions under varying $M$ values
and initial perturbations, together with the concentration-mass drift over the
complete rollout, are reported in
Figures~\ref{fig:spinodal_decomposition_test_final_shape_phi} and
\ref{fig:spinodal_decomposition_test_mass_drift}, respectively, in
Appendix~\ref{sec:spinodal-decomposition-additional-visualisations}.
Together, these results show that physics-informed fine-tuning
improves both predictive accuracy and physical consistency during long
autoregressive rollouts under complex nonlinear dynamics and stochastic
initial conditions.

\section{Concluding remarks}
\label{sec:conclusion}

This work presents PF-PINO, a physics-informed neural operator framework for parametric phase-field modelling of interfacial degradation and microstructural evolution. We systematically evaluate its performance across four representative benchmarks: pencil-electrode corrosion, electro-polishing corrosion, dendritic crystal solidification, and spinodal decomposition, which encompass a wide range of computational challenges, including sharp interface dynamics, strong anisotropy, non-periodic boundaries, out-of-distribution extrapolation, and complex pattern formation. The results consistently demonstrate that embedding PDE residuals into the FNO training loss yields substantial improvements in accuracy, generalisation, and long-term autoregressive stability compared to standard data-driven FNO. Quantitatively, PF-PINO reduces the relative $L^2$ error from \qty{1.58}{\percent} to \qty{0.53}{\percent} in the pencil-electrode corrosion benchmark and the relative Hausdorff distance from \num{4.81} to \num{1.10} in the dendritic solidification benchmark involving extrapolation beyond the training range.

These
improvements are consistent with the residual-based analysis in
Appendix~\ref{sec:residual-theory}.
Under local inverse stability,
reducing the PDE residual controls a contribution to the one-step prediction error
and constrains the learned solution graph near the physical solution set.
During autoregressive prediction,
smaller local update defects reduce their accumulated contribution
to the long-horizon error,
while rollout fine-tuning provides additional control on model-generated states.
For the periodic spinodal decomposition benchmark,
spectral differentiation weights the resolved Fourier modes
and helps suppress dynamically inconsistent spectral errors,
consistent with the improved structure factor.
Importantly, PF-PINO and FNO share identical network architectures.
Once the model parameters are fixed, a PF-PINO forward pass has the same
inference cost as an FNO forward pass. The additional evaluation of PDE
residuals is incurred during physics-informed training, while optional
test-time fine-tuning, when used, constitutes a separate optimisation stage.


Several avenues for future development are identified. Extension to three-dimensional phase-field simulations represents a natural and high-impact next step, where the resolution-invariant property of neural operators provides significant advantages over conventional solvers facing cubic scaling of spatial degrees of freedom. Fully data-free training strategies, in which network parameters are optimised exclusively from PDE residuals without reference simulation data, would further enhance data efficiency and broaden applicability to regimes where generating large high-fidelity datasets is prohibitively expensive. Integration with experimentally acquired microstructural data and structural reliability assessment frameworks could enable physics-consistent evaluation of material degradation under realistic operational conditions. Additionally, extension to coupled thermo-mechanical and electrochemical-mechanical problems would allow direct linking of microstructural evolution predictions to macroscopic structural response.

Looking forward, PF-PINO offers a promising pathway for engineering practice. Its capacity for accurate, parametric prediction across broad kinetic and morphological parameter spaces without retraining positions it as an efficient engine for high-throughput materials degradation assessment, microstructure-informed design optimisation, and structural reliability analysis---all contexts where large numbers of forward evaluations are routinely required. By embedding physical laws directly into the learning process, the framework provides a principled balance between the fidelity of phase-field modelling and the computational tractability demanded by engineering workflows, establishing a foundation for physics-consistent, data-efficient prediction of complex material evolution processes central to the structural integrity and long-term safety of engineering systems.

\section*{Author contributions}
Rujin Ma and Airong Chen conceptualized the research.
Nanxi Chen developed the methodology, conducted the investigation and formal analysis,
implemented the software, and wrote the original draft of the manuscript.
Rujin Ma and Airong Chen acquired funding and supervised the research.
Rujin Ma validated the results and reviewed and edited the manuscript.
All authors reviewed and approved the final version.

\section*{Declaration of competing interests}
The authors declare that they have no known competing financial interests or personal relationships that could have appeared to influence the work reported in this paper.

\section*{Data availability}
All code used in this study is released open source at \url{https://github.com/NanxiiChen/PF-PINO}, including data generation, neural network architectures, and training scripts for all benchmarks. 

\section*{Acknowledgements}
This work was supported by the National Natural Science Foundation of China (NSFC) under Grant No.\ 52478199 and No.\ 52238005. Nanxi Chen acknowledges support from the China Association for Science and Technology (CAST) through the Young Science and Technology Talent Cultivation Project (Doctoral Student Program). The valuable discussions with Dr.\ Chuanjie Cui (University of Oxford) are gratefully acknowledged.

\bibliographystyle{unsrt}  
\bibliography{bibfile,bibfile-no,misc}

@article{alhada-lahbabiMachineLearningSurrogate2023,
  title = {Machine {{Learning Surrogate Model}} for {{Acceleration}} of {{Ferroelectric Phase-Field Modeling}}},
  author = {{Alhada-Lahbabi}, K{\'e}vin and Deleruyelle, Damien and Gautier, Brice},
  year = 2023,
  month = jul,
  journal = {ACS Applied Electronic Materials},
  volume = {5},
  number = {7},
  pages = {3894--3907},
  publisher = {American Chemical Society},
  doi = {10.1021/acsaelm.3c00601},
  urldate = {2026-01-11},
  langid = {american}
}

@article{alhada-lahbabiMachineLearningSurrogate2024,
  title = {Machine Learning Surrogate for {{3D}} Phase-Field Modeling of Ferroelectric Tip-Induced Electrical Switching},
  author = {{Alhada--Lahbabi}, K{\'e}vin and Deleruyelle, Damien and Gautier, Brice},
  year = 2024,
  month = aug,
  journal = {npj Computational Materials},
  volume = {10},
  number = {1},
  pages = {197},
  publisher = {Nature Publishing Group},
  issn = {2057-3960},
  doi = {10.1038/s41524-024-01375-7},
  urldate = {2026-01-11},
  copyright = {2024 The Author(s)},
  langid = {english}
}

@article{azizzadenesheliNeuralOperatorsAccelerating2024,
  title = {Neural Operators for Accelerating Scientific Simulations and Design},
  author = {Azizzadenesheli, Kamyar and Kovachki, Nikola and Li, Zongyi and {Liu-Schiaffini}, Miguel and Kossaifi, Jean and Anandkumar, Anima},
  year = 2024,
  month = may,
  journal = {Nature Reviews Physics},
  volume = {6},
  number = {5},
  pages = {320--328},
  publisher = {Nature Publishing Group},
  issn = {2522-5820},
  doi = {10.1038/s42254-024-00712-5},
  urldate = {2025-10-30},
  copyright = {2024 Springer Nature Limited},
  langid = {english}
}

@article{bonnevilleAcceleratingPhaseField2025,
  title = {Accelerating Phase Field Simulations through a Hybrid Adaptive {{Fourier}} Neural Operator with {{U-net}} Backbone},
  author = {Bonneville, Christophe and Bieberdorf, Nathan and Hegde, Arun and Asta, Mark and Najm, Habib N. and Capolungo, Laurent and Safta, Cosmin},
  year = 2025,
  month = jan,
  journal = {npj Computational Materials},
  volume = {11},
  number = {1},
  pages = {14},
  publisher = {Nature Publishing Group},
  issn = {2057-3960},
  doi = {10.1038/s41524-024-01488-z},
  urldate = {2025-11-12},
  copyright = {2025 This is a U.S. Government work and not under copyright protection in the US; foreign copyright protection may apply},
  langid = {english}
}

@article{ciesielskiDeepOperatorNetwork2025,
  title = {Deep Operator Network Surrogate for Phase-Field Modeling of Metal Grain Growth during Solidification},
  author = {Ciesielski, Danielle and Li, Yulan and Hu, Shenyang and King, Ethan and Corbey, Jordan and Stinis, Panos},
  year = 2025,
  month = jan,
  journal = {Computational Materials Science},
  volume = {246},
  pages = {113417},
  issn = {0927-0256},
  doi = {10.1016/j.commatsci.2024.113417},
  urldate = {2025-12-14},
  langid = {american}
}

@misc{gangmeiLearningCoupledAllenCahn2025,
  title = {Learning Coupled {{Allen-Cahn}} and {{Cahn-Hilliard}} Phase-Field Equations Using {{Physics-Informed}} Neural Operator ({{PINO}})},
  author = {Gangmei, Gaijinliu and Rana, Santu and Rolfe, Bernard and Mitra, Kishalay and Bhattacharyya, Saswata},
  year = 2025,
  month = jul,
  eprint = {2507.18731},
  primaryclass = {cs.CE},
  doi = {10.48550/arXiv.2507.18731},
  url = {https://arxiv.org/abs/2507.18731},
  archiveprefix = {arXiv}
}

@article{xiangMorphologicalDesignTunable2024,
  title = {Morphological Design and Tunable Mechanical Properties of {{3D}} Spinodal Membrane Structures: Adaptive Coarse-Grained Modelling},
  author = {Xiang, Yujie and Tian, Jie and Tang, Keke and Wang, Xianqiao and Zhong, Zheng},
  journal = {Acta Mechanica Sinica},
  year = 2024,
  volume = {40},
  number = {8},
  pages = {423655},
  doi = {10.1007/s10409-024-23655-x}
}

@article{gaoCNNBasedSurrogatePhase2023,
  title = {{{CNN-Based Surrogate}} for the {{Phase Field Damage Model}}: {{Generalization}} across {{Microstructure Parameters}} for {{Composite Materials}}},
  shorttitle = {{{CNN-Based Surrogate}} for the {{Phase Field Damage Model}}},
  author = {Gao, Yuxiang and Berger, Matthew and Duddu, Ravindra},
  year = 2023,
  month = jun,
  journal = {Journal of Engineering Mechanics},
  volume = {149},
  number = {6},
  pages = {04023025},
  issn = {0733-9399, 1943-7889},
  doi = {10.1061/JENMDT.EMENG-6936},
  urldate = {2026-01-11},
  langid = {english}
}

@article{huAcceleratingPhasefieldPredictions2022,
  title = {Accelerating Phase-Field Predictions via Recurrent Neural Networks Learning the Microstructure Evolution in Latent Space},
  author = {Hu, C. and Martin, S. and Dingreville, R.},
  year = 2022,
  month = jul,
  journal = {Computer Methods in Applied Mechanics and Engineering},
  volume = {397},
  pages = {115128},
  issn = {0045-7825},
  doi = {10.1016/j.cma.2022.115128},
  urldate = {2026-01-07},
  langid = {american}
}

@article{huAcceleratingPhasefieldSimulation2025,
  title = {Accelerating Phase-Field Simulation of Coupled Microstructural Evolution Using Autoencoder-Based Recurrent Neural Networks},
  author = {Gesch, Aidan and Hu, Chongze},
  year = 2025,
  month = jun,
  journal = {Journal of Materials Informatics},
  volume = {5},
  number = {4},
  pages = {42},
  publisher = {OAE Publishing Inc.},
  issn = {2770-372X},
  doi = {10.20517/jmi.2025.23},
  url = {https://www.oaepublish.com/articles/jmi.2025.23},
  urldate = {2026-01-11},
  langid = {english}
}

@article{karmaPhasefieldModelDendritic1999,
  title = {Phase-Field Model of Dendritic Sidebranching with Thermal Noise},
  author = {Karma, Alain and Rappel, Wouter-Jan},
  year = 1999,
  month = oct,
  journal = {Physical Review E},
  volume = {60},
  number = {4},
  pages = {3614--3625},
  issn = {1063-651X, 1095-3787},
  doi = {10.1103/PhysRevE.60.3614},
  urldate = {2025-11-24},
  copyright = {http://link.aps.org/licenses/aps-default-license},
  langid = {english}
}

@article{karmaQuantitativePhasefieldModeling1998,
  title = {Quantitative Phase-Field Modeling of Dendritic Growth in Two and Three Dimensions},
  author = {Karma, Alain and Rappel, Wouter-Jan},
  year = 1998,
  month = apr,
  journal = {Physical Review E},
  volume = {57},
  number = {4},
  pages = {4323--4349},
  issn = {1063-651X},
  doi = {10.1103/PhysRevE.57.4323},
  langid = {english}
}

@article{kovachkiNeuralOperatorLearning2024,
  title = {Neural {{Operator}}: {{Learning Maps Between Function Spaces With Applications to PDEs}}},
  shorttitle = {Neural {{Operator}}},
  author = {Kovachki, Nikola and Li, Zongyi and Liu, Burigede and Azizzadenesheli, Kamyar and Bhattacharya, Kaushik and Stuart, Andrew and Anandkumar, Anima},
  year = 2023,
  journal = {Journal of Machine Learning Research},
  volume = {24},
  number = {89},
  pages = {1--97},
  url = {https://www.jmlr.org/papers/v24/21-1524.html},
  urldate = {2025-10-27},
  langid = {american}
}

@article{krekhovPhaseSeparationPresence2004,
  title = {Phase Separation in the Presence of Spatially Periodic Forcing},
  author = {Krekhov, A. P. and Kramer, L.},
  year = 2004,
  month = dec,
  journal = {Physical Review E},
  volume = {70},
  number = {6},
  pages = {061801},
  issn = {1539-3755, 1550-2376},
  doi = {10.1103/PhysRevE.70.061801},
  urldate = {2025-12-06},
  copyright = {http://link.aps.org/licenses/aps-default-license},
  langid = {english}
}

@inproceedings{liFourierNeuralOperator2021,
  title = {Fourier Neural Operator for Parametric Partial Differential Equations},
  author = {Li, Zongyi and Kovachki, Nikola and Azizzadenesheli, Kamyar and Liu, Burigede and Bhattacharya, Kaushik and Stuart, Andrew and Anandkumar, Anima},
  year = 2021,
  booktitle = {International Conference on Learning Representations},
  eprint = {2010.08895},
  primaryclass = {cs.LG},
  url = {https://openreview.net/forum?id=c8P9NQVtmnO},
  urldate = {2024-01-20},
  archiveprefix = {arXiv},
  langid = {american}
}

@article{liPhasefieldDeepONetPhysicsinformed2023,
  title = {Phase-Field {{DeepONet}}: Physics-Informed Deep Operator Neural Network for Fast Simulations of Pattern Formation Governed by Gradient Flows of Free-Energy Functionals},
  shorttitle = {Phase-{{Field DeepONet}}},
  author = {Li, Wei and Bazant, Martin Z. and Zhu, Juner},
  year = 2023,
  month = nov,
  journal = {Computer Methods in Applied Mechanics and Engineering},
  volume = {416},
  pages = {116299},
  issn = {00457825},
  doi = {10.1016/j.cma.2023.116299},
  urldate = {2024-02-28},
  langid = {english}
}

@article{liPhysicsInformedNeuralOperator2023,
  title = {Physics-{{Informed Neural Operator}} for {{Learning Partial Differential Equations}}},
  author = {Li, Zongyi and Zheng, Hongkai and Kovachki, Nikola and Jin, David and Chen, Haoxuan and Liu, Burigede and Azizzadenesheli, Kamyar and Anandkumar, Anima},
  year = 2024,
  month = may,
  journal = {ACM/IMS Journal of Data Science},
  volume = {1},
  number = {3},
  pages = {1--27},
  issn = {2831-3194},
  doi = {10.1145/3648506},
  url = {https://doi.org/10.1145/3648506},
  urldate = {2025-10-26}
}

@article{liReviewApplicationsPhase2017a,
  title = {A Review: Applications of the Phase Field Method in Predicting Microstructure and Property Evolution of Irradiated Nuclear Materials},
  shorttitle = {A Review},
  author = {Li, Yulan and Hu, Shenyang and Sun, Xin and Stan, Marius},
  year = 2017,
  month = apr,
  journal = {npj Computational Materials},
  volume = {3},
  number = {1},
  pages = {16},
  issn = {2057-3960},
  doi = {10.1038/s41524-017-0018-y},
  urldate = {2026-01-11},
  langid = {english}
}

@article{luComprehensiveFairComparison2022,
  title = {A Comprehensive and Fair Comparison of Two Neural Operators (with Practical Extensions) Based on {{FAIR}} Data},
  author = {Lu, Lu and Meng, Xuhui and Cai, Shengze and Mao, Zhiping and Goswami, Somdatta and Zhang, Zhongqiang and Karniadakis, George Em},
  year = 2022,
  month = apr,
  journal = {Computer Methods in Applied Mechanics and Engineering},
  volume = {393},
  pages = {114778},
  issn = {00457825},
  doi = {10.1016/j.cma.2022.114778},
  urldate = {2025-10-30},
  langid = {english}
}

@article{luLearningNonlinearOperators2021,
  title = {Learning Nonlinear Operators via {{DeepONet}} Based on the Universal Approximation Theorem of Operators},
  author = {Lu, Lu and Jin, Pengzhan and Pang, Guofei and Zhang, Zhongqiang and Karniadakis, George Em},
  year = 2021,
  month = mar,
  journal = {Nature Machine Intelligence},
  volume = {3},
  number = {3},
  pages = {218--229},
  issn = {2522-5839},
  doi = {10.1038/s42256-021-00302-5},
  urldate = {2024-01-18},
  langid = {english}
}

@article{mccabeStabilityAutoregressiveNeural2023,
  title = {Towards {{Stability}} of {{Autoregressive Neural Operators}}},
  author = {McCabe, Michael and Harrington, Peter and Subramanian, Shashank and Brown, Jed},
  year = 2023,
  journal = {Transactions on Machine Learning Research},
  issn = {2835-8856},
  url = {https://openreview.net/forum?id=RFfUUtKYOG},
  urldate = {2025-11-04},
  langid = {english}
}

@article{montesdeocazapiainAcceleratingPhasefieldbasedMicrostructure2021,
  title = {Accelerating Phase-Field-Based Microstructure Evolution Predictions via Surrogate Models Trained by Machine Learning Methods},
  author = {{Montes de Oca Zapiain}, David and Stewart, James A. and Dingreville, R{\'e}mi},
  year = 2021,
  month = jan,
  journal = {npj Computational Materials},
  volume = {7},
  number = {1},
  pages = {3},
  publisher = {Nature Publishing Group},
  issn = {2057-3960},
  doi = {10.1038/s41524-020-00471-8},
  urldate = {2026-01-11},
  copyright = {2021 This is a U.S. Government work and not under copyright protection in the US; foreign copyright protection may apply},
  langid = {english}
}

@article{onetoInformedMachineLearning2025,
  title = {Informed {{Machine Learning}}: {{Excess}} Risk and Generalization},
  shorttitle = {Informed {{Machine Learning}}},
  author = {Oneto, Luca and Ridella, Sandro and Anguita, Davide},
  year = 2025,
  month = sep,
  journal = {Neurocomputing},
  volume = {646},
  pages = {130521},
  issn = {0925-2312},
  doi = {10.1016/j.neucom.2025.130521},
  urldate = {2026-01-11}
}

@article{oommenLearningTwophaseMicrostructure2022,
  title = {Learning Two-Phase Microstructure Evolution Using Neural Operators and Autoencoder Architectures},
  author = {Oommen, Vivek and Shukla, Khemraj and Goswami, Somdatta and Dingreville, R{\'e}mi and Karniadakis, George Em},
  year = 2022,
  month = sep,
  journal = {npj Computational Materials},
  volume = {8},
  number = {1},
  pages = {190},
  publisher = {Nature Publishing Group},
  issn = {2057-3960},
  doi = {10.1038/s41524-022-00876-7},
  urldate = {2025-12-16},
  copyright = {2022 The Author(s)},
  langid = {english}
}

@article{peivasteMachinelearningbasedSurrogateModeling2022,
  title = {Machine-Learning-Based Surrogate Modeling of Microstructure Evolution Using Phase-Field},
  author = {Peivaste, Iman and Siboni, Nima H. and Alahyarizadeh, Ghasem and Ghaderi, Reza and Svendsen, Bob and Raabe, Dierk and Mianroodi, Jaber Rezaei},
  year = 2022,
  month = nov,
  journal = {Computational Materials Science},
  volume = {214},
  pages = {111750},
  issn = {0927-0256},
  doi = {10.1016/j.commatsci.2022.111750},
  urldate = {2026-01-11},
  langid = {american}
}

@article{rabehBenchmarkingScientificMachinelearning2025,
  title = {Benchmarking Scientific Machine-Learning Approaches for Flow Prediction around Complex Geometries},
  author = {Rabeh, Ali and Herron, Ethan and Balu, Aditya and Sarkar, Soumik and Hegde, Chinmay and Krishnamurthy, Adarsh and Ganapathysubramanian, Baskar},
  year = 2025,
  month = oct,
  journal = {Communications Engineering},
  volume = {4},
  number = {1},
  pages = {182},
  publisher = {Nature Publishing Group},
  issn = {2731-3395},
  doi = {10.1038/s44172-025-00513-3},
  urldate = {2026-01-11},
  copyright = {2025 The Author(s)},
  langid = {english}
}

@article{ramanarayanPhaseFieldStudy2003,
  title = {Phase Field Study of Grain Boundary Effects on Spinodal Decomposition},
  author = {Ramanarayan, H. and Abinandanan, T. A.},
  year = 2003,
  month = sep,
  journal = {Acta Materialia},
  volume = {51},
  number = {16},
  pages = {4761--4772},
  issn = {1359-6454},
  doi = {10.1016/S1359-6454(03)00301-X},
  urldate = {2025-12-09}
}

@article{rosofskyApplicationsPhysicsInformed2023,
  title = {Applications of Physics Informed Neural Operators},
  author = {Rosofsky, Shawn G. and Majed, Hani Al and Huerta, E. A.},
  year = 2023,
  month = may,
  journal = {Machine Learning: Science and Technology},
  volume = {4},
  number = {2},
  pages = {025022},
  publisher = {IOP Publishing},
  issn = {2632-2153},
  doi = {10.1088/2632-2153/acd168},
  urldate = {2026-01-11},
  langid = {english}
}

@article{tiwariTimeSeriesForecasting2025,
  title = {Time Series Forecasting of Multiphase Microstructure Evolution Using Deep Learning},
  author = {Tiwari, Saurabh and Satpute, Prathamesh and Ghosh, Supriyo},
  year = 2025,
  month = jan,
  journal = {Computational Materials Science},
  volume = {247},
  pages = {113518},
  issn = {0927-0256},
  doi = {10.1016/j.commatsci.2024.113518},
  urldate = {2026-01-11}
}

@article{valentePhysicsconsistentMachineLearning2025,
  title = {Physics-Consistent Machine Learning: Output Projection onto Physical Manifolds},
  shorttitle = {Physics-Consistent Machine Learning},
  author = {Valente, Matilde and Dias, Tiago C. and Guerra, Vasco and Ventura, Rodrigo},
  year = 2025,
  month = nov,
  journal = {Communications Physics},
  volume = {8},
  number = {1},
  eprint = {2502.15755},
  primaryclass = {cs},
  pages = {433},
  issn = {2399-3650},
  doi = {10.1038/s42005-025-02329-1},
  urldate = {2026-01-11},
  archiveprefix = {arXiv}
}

@article{wangLearningSolutionOperator2021,
  title = {Learning the Solution Operator of Parametric Partial Differential Equations with Physics-Informed {{DeepONets}}},
  author = {Wang, Sifan and Wang, Hanwen and Perdikaris, Paris},
  year = 2021,
  month = sep,
  journal = {Science Advances},
  volume = {7},
  number = {40},
  pages = {eabi8605},
  publisher = {American Association for the Advancement of Science},
  doi = {10.1126/sciadv.abi8605},
  urldate = {2024-01-21},
  langid = {american}
}

@article{wuSimGateDeepLearning2025,
  title = {{{SimGate}}: {{A}} Deep Learning Surrogate Model for Predicting Microstructure Evolution Using the Phase-Field Method},
  shorttitle = {{{SimGate}}},
  author = {Wu, Pin and Huang, Haiwang and Yang, Qingcheng and Qian, Bo and Gao, Yongxin and Yang, Yiguo and Zhang, Huiran and Zhen, Qiang},
  year = 2025,
  month = jun,
  journal = {Computational Materials Science},
  volume = {256},
  pages = {113883},
  issn = {0927-0256},
  doi = {10.1016/j.commatsci.2025.113883},
  urldate = {2026-01-11},
  langid = {american}
}

@misc{xueEquivariantUShapedNeural2025,
  title = {Equivariant {{U-Shaped Neural Operators}} for the {{Cahn-Hilliard Phase-Field Model}}},
  author = {Xue, Xiao and ten Eikelder, Marco F. P. and Yang, Tianyue and Li, Yiqing and He, Kan and Wang, Shuo and Coveney, Peter V.},
  year = 2025,
  month = sep,
  number = {arXiv:2509.01293},
  eprint = {2509.01293},
  primaryclass = {cs},
  publisher = {arXiv},
  doi = {10.48550/arXiv.2509.01293},
  urldate = {2025-12-14},
  archiveprefix = {arXiv},
  langid = {american}
}

@article{yangEfficientLinearStabilized2019,
  title = {Efficient Linear, Stabilized, Second-Order Time Marching Schemes for an Anisotropic Phase Field Dendritic Crystal Growth Model},
  author = {Yang, Xiaofeng},
  year = 2019,
  month = apr,
  journal = {Computer Methods in Applied Mechanics and Engineering},
  volume = {347},
  pages = {316--339},
  issn = {0045-7825},
  doi = {10.1016/j.cma.2018.12.012},
  urldate = {2025-11-23},
  langid = {english}
}

@article{zhangNumericalSolutionPhasefield2023,
  title = {Numerical Solution to Phase-Field Model of Solidification: {{A}} Review},
  shorttitle = {Numerical Solution to Phase-Field Model of Solidification},
  author = {Zhang, Ang and Guo, Zhipeng and Jiang, Bin and Xiong, Shoumei and Pan, Fusheng},
  year = 2023,
  month = sep,
  journal = {Computational Materials Science},
  volume = {228},
  pages = {112366},
  issn = {0927-0256},
  doi = {10.1016/j.commatsci.2023.112366},
  urldate = {2026-01-11},
  langid = {american}
}

@article{berroneVariationalPhysicsInformed2022,
  title = {Variational Physics Informed Neural Networks: The Role of Quadratures and Test Functions},
  author = {Berrone, Stefano and Canuto, Claudio and Pintore, Moreno},
  year = 2022,
  journal = {Journal of Scientific Computing},
  volume = {92},
  number = {3},
  pages = {100},
  issn = {1573-7691},
  doi = {10.1007/s10915-022-01950-4}
}

@article{boettingerPhasefieldSimulationSolidification2002,
  title = {Phase-Field Simulation of Solidification},
  author = {Boettinger, W. J. and Warren, J. A. and Beckermann, C. and Karma, A.},
  year = 2002,
  month = aug,
  journal = {Annual Review of Materials Research},
  volume = {32},
  number = {1},
  pages = {163--194},
  issn = {1531-7331, 1545-4118},
  doi = {10.1146/annurev.matsci.32.101901.155803},
  urldate = {2023-11-28},
  langid = {english}
}

@inproceedings{chenGradNormGradientNormalization2018,
  title = {{{GradNorm}}: {{Gradient Normalization}} for {{Adaptive Loss Balancing}} in {{Deep Multitask Networks}}},
  shorttitle = {{{GradNorm}}},
  booktitle = {Proceedings of the 35th {{International Conference}} on {{Machine Learning}}},
  author = {Chen, Zhao and Badrinarayanan, Vijay and Lee, Chen-Yu and Rabinovich, Andrew},
  editor = {Dy, Jennifer and Krause, Andreas},
  year = 2018,
  month = jul,
  series = {Proceedings of Machine Learning Research},
  volume = {80},
  pages = {794--803},
  publisher = {PMLR},
  issn = {2640-3498},
  url = {https://proceedings.mlr.press/v80/chen18a.html},
  urldate = {2024-12-29},
  langid = {english}
}

@article{chenPFPINNsPhysicsinformedNeural2025b,
  title = {{{PF-PINNs}}: {{Physics-informed}} Neural Networks for Solving Coupled {{Allen-Cahn}} and {{Cahn-Hilliard}} Phase Field Equations},
  author = {Chen, Nanxi and Lucarini, Sergio and Ma, Rujin and Chen, Airong and Cui, Chuanjie},
  year = 2025,
  month = may,
  journal = {Journal of Computational Physics},
  volume = {529},
  pages = {113843},
  issn = {0021-9991},
  doi = {10.1016/j.jcp.2025.113843},
  url = {https://doi.org/10.1016/j.jcp.2025.113843}
}

@article{chenPhasefieldModelsMicrostructure2002,
  title = {Phase-Field Models for Microstructure Evolution},
  author = {Chen, Long-Qing},
  year = 2002,
  month = aug,
  journal = {Annual Review of Materials Research},
  volume = {32},
  number = {1},
  pages = {113--140},
  issn = {1531-7331, 1545-4118},
  doi = {10.1146/annurev.matsci.32.112001.132041},
  urldate = {2024-04-29},
  langid = {english}
}

@article{chenSharpPINNsStaggeredHardconstrained2025,
  title = {Sharp-{{PINNs}}: {{Staggered}} Hard-Constrained Physics-Informed Neural Networks for Phase Field Modelling of Corrosion},
  author = {Chen, Nanxi and Cui, Chuanjie and Ma, Rujin and Chen, Airong and Wang, Sifan},
  year = 2025,
  journal = {Computer Methods in Applied Mechanics and Engineering},
  volume = {447},
  pages = {118346},
  issn = {0045-7825},
  doi = {10.1016/j.cma.2025.118346},
  langid = {american}
}

@article{cuiGeneralisedMultiphasefieldTheory2022b,
  title = {A Generalised, Multi-Phase-Field Theory for Dissolution-Driven Stress Corrosion Cracking and Hydrogen Embrittlement},
  author = {Cui, Chuanjie and Ma, Rujin and {Mart{\'i}nez-Pa{\~n}eda}, Emilio},
  year = 2022,
  month = sep,
  journal = {Journal of the Mechanics and Physics of Solids},
  volume = {166},
  pages = {104951},
  issn = {00225096},
  doi = {10.1016/j.jmps.2022.104951},
  urldate = {2024-05-09},
  langid = {english}
}

@article{cuiPhaseFieldFormulation2021,
  title = {A Phase Field Formulation for Dissolution-Driven Stress Corrosion Cracking},
  author = {Cui, Chuanjie and Ma, Rujin and {Mart{\'i}nez-Pa{\~n}eda}, Emilio},
  year = 2021,
  month = feb,
  journal = {Journal of the Mechanics and Physics of Solids},
  volume = {147},
  pages = {104254},
  issn = {00225096},
  doi = {10.1016/j.jmps.2020.104254},
  urldate = {2023-10-13},
  langid = {english}
}

@article{cuomoScientificMachineLearning2022,
  title = {Scientific Machine Learning through Physics-Informed Neural Networks: Where We Are and What's Next},
  shorttitle = {Scientific {{Machine Learning}} through {{Physics-Informed Neural Networks}}},
  author = {Cuomo, Salvatore and {Di Cola}, Vincenzo Schiano and Giampaolo, Fabio and Rozza, Gianluigi and Raissi, Maziar and Piccialli, Francesco},
  year = 2022,
  month = jul,
  journal = {Journal of Scientific Computing},
  volume = {92},
  number = {3},
  pages = {88},
  doi = {10.1007/s10915-022-01939-z},
  url = {https://doi.org/10.1007/s10915-022-01939-z},
  urldate = {2022-11-30}
}

@article{deryckErrorAnalysisPhysics2021,
  title = {Error Analysis for Physics-Informed Neural Networks ({{PINNs}}) Approximating {{Kolmogorov PDEs}}},
  author = {De Ryck, Tim and Mishra, Siddhartha},
  year = 2022,
  month = nov,
  journal = {Advances in Computational Mathematics},
  volume = {48},
  number = {6},
  pages = {79},
  doi = {10.1007/s10444-022-09985-9},
  url = {https://doi.org/10.1007/s10444-022-09985-9},
  urldate = {2023-06-01}
}

@article{huttenlocherComparingImagesHausdorff1993,
  title = {Comparing Images Using the {Hausdorff} Distance},
  author = {Huttenlocher, Daniel P. and Klanderman, Gregory A. and Rucklidge, William J.},
  year = 1993,
  journal = {IEEE Transactions on Pattern Analysis and Machine Intelligence},
  volume = {15},
  number = {9},
  pages = {850--863},
  doi = {10.1109/34.232073},
  url = {https://doi.org/10.1109/34.232073}
}

@article{karniadakisPhysicsinformedMachineLearning2021,
  title = {Physics-Informed Machine Learning},
  author = {Karniadakis, George Em and Kevrekidis, Ioannis G. and Lu, Lu and Perdikaris, Paris and Wang, Sifan and Yang, Liu},
  year = 2021,
  month = jun,
  journal = {Nature Reviews Physics},
  volume = {3},
  number = {6},
  pages = {422--440},
  publisher = {Nature Publishing Group},
  issn = {2522-5820},
  doi = {10.1038/s42254-021-00314-5},
  urldate = {2023-11-29},
  copyright = {2021 Springer Nature Limited},
  langid = {english}
}

@article{kharazmiHpVPINNsVariationalPhysicsinformed2021,
  title = {{hp}-{{VPINNs}}: {{Variational}} Physics-Informed Neural Networks with Domain Decomposition},
  shorttitle = {{hp}-{{VPINNs}}},
  author = {Kharazmi, Ehsan and Zhang, Zhongqiang and Karniadakis, George E. M.},
  year = 2021,
  month = feb,
  journal = {Computer Methods in Applied Mechanics and Engineering},
  volume = {374},
  pages = {113547},
  issn = {0045-7825},
  doi = {10.1016/j.cma.2020.113547},
  urldate = {2025-12-19}
}

@article{kimPhasefieldModelBinary1999,
  title = {Phase-Field Model for Binary Alloys},
  author = {Kim, Seong Gyoon and Kim, Won Tae and Suzuki, Toshio},
  year = 1999,
  month = dec,
  journal = {Physical Review E},
  volume = {60},
  number = {6},
  pages = {7186--7197},
  issn = {1063-651X, 1095-3787},
  doi = {10.1103/PhysRevE.60.7186},
  urldate = {2023-10-12},
  langid = {english}
}

@article{liProbabilisticPhysicsinformedNeural2024,
  title = {Probabilistic Physics-Informed Neural Network for Seismic Petrophysical Inversion},
  author = {Li, Peng and Liu, Mingliang and Alfarraj, Motaz and Tahmasebi, Pejman and Grana, Dario},
  year = 2024,
  month = mar,
  journal = {GEOPHYSICS},
  volume = {89},
  number = {2},
  pages = {M17-M32},
  issn = {0016-8033, 1942-2156},
  doi = {10.1190/geo2023-0214.1},
  urldate = {2025-01-17},
  langid = {english}
}

@article{liTutorialsPhysicsinformedMachine2024,
  title = {Tutorials: Physics-Informed Machine Learning Methods of Computing {{1D}} Phase-Field Models},
  shorttitle = {Tutorials},
  author = {Li, Wei and Fang, Ruqing and Jiao, Junning and Vassilakis, Georgios N. and Zhu, Juner},
  year = 2024,
  month = aug,
  journal = {APL Machine Learning},
  volume = {2},
  number = {3},
  pages = {031101},
  issn = {2770-9019},
  doi = {10.1063/5.0205159},
  urldate = {2024-12-08}
}

@article{maiPhaseFieldModel2016,
  title = {A Phase Field Model for Simulating the Pitting Corrosion},
  author = {Mai, Weijie and Soghrati, Soheil and Buchheit, Rudolph G.},
  year = 2016,
  month = sep,
  journal = {Corrosion Science},
  volume = {110},
  pages = {157--166},
  issn = {0010-938X},
  doi = {10.1016/j.corsci.2016.04.001},
  urldate = {2023-10-09},
  langid = {english}
}

@article{matteyNovelSequentialMethod2022,
  title = {A Novel Sequential Method to Train Physics Informed Neural Networks for Allen Cahn and Cahn Hilliard Equations},
  author = {Mattey, Revanth and Ghosh, Susanta},
  year = 2022,
  month = feb,
  journal = {Computer Methods in Applied Mechanics and Engineering},
  volume = {390},
  pages = {114474},
  issn = {0045-7825},
  doi = {10.1016/j.cma.2021.114474},
  urldate = {2023-09-27}
}

@article{nabianEfficientTrainingPhysicsinformed2021,
  title = {Efficient Training of Physics-Informed Neural Networks via Importance Sampling},
  author = {Nabian, Mohammad Amin and Gladstone, Rini Jasmine and Meidani, Hadi},
  year = 2021,
  journal = {Computer-Aided Civil and Infrastructure Engineering},
  volume = {36},
  number = {8},
  pages = {962--977},
  issn = {1467-8667},
  doi = {10.1111/mice.12685},
  urldate = {2023-09-27},
  copyright = {\copyright{} 2021 Computer-Aided Civil and Infrastructure Engineering},
  langid = {english}
}

@article{qiuPhysicsinformedNeuralNetworks2022,
  title = {Physics-Informed Neural Networks for Phase-Field Method in Two-Phase Flow},
  author = {Qiu, Rundi and Huang, Renfang and Xiao, Yao and Wang, Jingzhu and Zhang, Zhen and Yue, Jieshun and Zeng, Zhong and Wang, Yiwei},
  year = 2022,
  month = may,
  journal = {Physics of Fluids},
  volume = {34},
  number = {5},
  pages = {052109},
  publisher = {American Institute of Physics},
  issn = {1070-6631},
  doi = {10.1063/5.0091063},
  urldate = {2022-11-30}
}

@article{raissiPhysicsinformedNeuralNetworks2019,
  title = {Physics-Informed Neural Networks: A Deep Learning Framework for Solving Forward and Inverse Problems Involving Nonlinear Partial Differential Equations},
  shorttitle = {Physics-Informed Neural Networks},
  author = {Raissi, M. and Perdikaris, P. and Karniadakis, G. E.},
  year = 2019,
  month = feb,
  journal = {Journal of Computational Physics},
  volume = {378},
  pages = {686--707},
  issn = {0021-9991},
  doi = {10.1016/j.jcp.2018.10.045},
  urldate = {2022-09-24},
  langid = {english}
}

@article{steinbachPhasefieldModelsMaterials2009,
  title = {Phase-Field Models in Materials Science},
  author = {Steinbach, Ingo},
  year = 2009,
  month = jul,
  journal = {Modelling and Simulation in Materials Science and Engineering},
  volume = {17},
  number = {7},
  pages = {073001},
  issn = {0965-0393},
  doi = {10.1088/0965-0393/17/7/073001},
  urldate = {2023-11-26},
  langid = {english}
}

@article{tahaMetricsEvaluating3D2015,
  title = {Metrics for Evaluating {3D} Medical Image Segmentation: Analysis, Selection, and Tool},
  author = {Taha, Abdel Aziz and Hanbury, Allan},
  year = 2015,
  month = aug,
  journal = {BMC Medical Imaging},
  volume = {15},
  number = {1},
  pages = {29},
  doi = {10.1186/s12880-015-0068-x},
  url = {https://doi.org/10.1186/s12880-015-0068-x}
}

@misc{wangExpertsGuideTraining2023,
  title = {An Expert's Guide to Training Physics-Informed Neural Networks},
  author = {Wang, Sifan and Sankaran, Shyam and Wang, Hanwen and Perdikaris, Paris},
  year = 2023,
  month = aug,
  number = {arXiv:2308.08468},
  eprint = {2308.08468},
  primaryclass = {physics},
  publisher = {arXiv},
  doi = {10.48550/arXiv.2308.08468},
  urldate = {2024-06-03},
  archiveprefix = {arXiv}
}

@misc{wangGradientAlignmentPhysicsinformed2025a,
  title = {Gradient {{Alignment}} in {{Physics-informed Neural Networks}}: {{A Second-Order Optimization Perspective}}},
  shorttitle = {Gradient {{Alignment}} in {{Physics-informed Neural Networks}}},
  author = {Wang, Sifan and Bhartari, Ananyae Kumar and Li, Bowen and Perdikaris, Paris},
  year = 2025,
  month = feb,
  number = {arXiv:2502.00604},
  eprint = {2502.00604},
  primaryclass = {cs},
  publisher = {arXiv},
  doi = {10.48550/arXiv.2502.00604},
  urldate = {2025-02-15},
  archiveprefix = {arXiv}
}

@article{wight2020solving,
  title = {Solving Allen-Cahn and Cahn-Hilliard Equations Using the Adaptive Physics Informed Neural Networks},
  author = {Wight, Colby L and Zhao, Jia},
  year = 2021,
  month = jun,
  journal = {Communications in Computational Physics},
  volume = {29},
  number = {3},
  pages = {930--954},
  issn = {1815-2406, 1991-7120},
  doi = {10.4208/cicp.OA-2020-0086},
  urldate = {2022-10-11}
}

@article{vedantam2006efficient,
  title = {Efficient numerical algorithm for multiphase field simulations},
  author = {Vedantam, Srikanth and Patnaik, B. S. V.},
  journal = {Physical Review E},
  volume = {73},
  number = {1},
  pages = {016703},
  numpages = {8},
  year = {2006},
  month = {Jan},
  publisher = {American Physical Society},
  doi = {10.1103/PhysRevE.73.016703},
  url = {https://link.aps.org/doi/10.1103/PhysRevE.73.016703}
}

@article{manav2024phase,
title = {Phase-field modeling of fracture with physics-informed deep learning},
journal = {Computer Methods in Applied Mechanics and Engineering},
volume = {429},
pages = {117104},
year = {2024},
issn = {0045-7825},
doi = {https://doi.org/10.1016/j.cma.2024.117104},
url = {https://www.sciencedirect.com/science/article/pii/S0045782524003608},
author = {M. Manav and R. Molinaro and S. Mishra and L. {De Lorenzis}}
}

@article{chen2025enforcinghiddenphysicsphysicsinformed,
  title = {Enforcing Hidden Physics in Physics-Informed Neural Networks},
  author = {Chen, Nanxi and Wang, Sifan and Ma, Rujin and Chen, Airong and Cui, Chuanjie},
  year = 2026,
  month = jul,
  journal = {Communications Physics},
  publisher = {Nature Publishing Group},
  issn = {2399-3650},
  doi = {10.1038/s42005-026-02743-z},
  urldate = {2026-07-24},
  copyright = {2026 The Author(s)},
  langid = {english},
}

@article{landolt2003electrochemical,
title = {Electrochemical micromachining, polishing and surface structuring of metals: fundamental aspects and new developments},
journal = {Electrochimica Acta},
volume = {48},
number = {20},
pages = {3185-3201},
year = {2003},
note = {Electrochemistry in Molecular and Microscopic Dimensions},
issn = {0013-4686},
doi = {https://doi.org/10.1016/S0013-4686(03)00368-2},
url = {https://www.sciencedirect.com/science/article/pii/S0013468603003682},
author = {D. Landolt and P.-F. Chauvy and O. Zinger}
}

@article{cook1970brownian,
  title = {Brownian motion in spinodal decomposition},
  author = {Cook, H. E.},
  year = {1970},
  month = mar,
  journal = {Acta Metallurgica},
  volume = {18},
  number = {3},
  pages = {297--306},
  issn = {0001-6160},
  doi = {10.1016/0001-6160(70)90144-6},
  url = {https://www.sciencedirect.com/science/article/pii/0001616070901446}
}

@article{wei2016origin,
  title = {Origin of grain orientation during solidification of an aluminum alloy},
  author = {Wei, H. L. and Elmer, J. W. and DebRoy, T.},
  year = {2016},
  month = aug,
  journal = {Acta Materialia},
  volume = {115},
  pages = {123--131},
  publisher = {Elsevier},
  doi = {10.1016/j.actamat.2016.05.057},
  url = {https://doi.org/10.1016/j.actamat.2016.05.057}
}

@article{quested2005grain,
  title = {Grain refinement of {Al} alloys: Mechanisms determining as-cast grain size in directional solidification},
  author = {Quested, T. E. and Greer, A. L.},
  year = {2005},
  month = oct,
  journal = {Acta Materialia},
  volume = {53},
  number = {17},
  pages = {4643--4653},
  publisher = {Elsevier},
  doi = {10.1016/j.actamat.2005.06.018},
  url = {https://doi.org/10.1016/j.actamat.2005.06.018}
}

@article{schmid2011thermodynamic,
  title = {Thermodynamic aspects of grain growth restriction in multicomponent alloy solidification},
  author = {Schmid-Fetzer, R. and Kozlov, A.},
  year = {2011},
  month = sep,
  journal = {Acta Materialia},
  volume = {59},
  number = {15},
  pages = {6133--6144},
  publisher = {Elsevier},
  doi = {10.1016/j.actamat.2011.06.026},
  url = {https://doi.org/10.1016/j.actamat.2011.06.026}
}

@article{ditchek1979applications,
  title = {Applications of Spinodal Alloys},
  author = {Ditchek, B. and Schwartz, L. H.},
  year = {1979},
  month = aug,
  journal = {Annual Review of Materials Science},
  volume = {9},
  number = {1},
  pages = {219--253},
  publisher = {Annual Reviews},
  doi = {10.1146/annurev.ms.09.080179.001251},
  url = {https://doi.org/10.1146/annurev.ms.09.080179.001251}
}

\appendix

\setcounter{figure}{0}
\setcounter{table}{0}

\renewcommand{\thetable}{A\arabic{table}}
\renewcommand{\thefigure}{A\arabic{figure}}

\section{Numerical derivation of physics-informed neural operators (PINOs)}
\label{sec:numerical-derivation-pino}

Physical constraints imposed by the governing PDEs are incorporated into the NO framework via a residual loss term. Here, $\bm{u}(\bm{x},t)$ denotes the complete multi-channel solution field, while $u_q(\bm{x},t)$ denotes its $q$-th scalar component. For a typical one-step prediction from time $t_n$ to $t_{n+1}$, the temporal derivative of the solution field is approximated using a first-order forward difference scheme:
\begin{equation}
    \frac{\partial \bm{u}(\bm{x},t)}{\partial t} \approx \frac{\bm{u}(\bm{x}, t_{n+1}) - \bm{u}(\bm{x}, t_n)}{\Delta t},
\end{equation}
where $\Delta t = t_{n+1} - t_n$ denotes the time step. Spatial derivatives are computed using either finite difference schemes or spectral methods, depending on the boundary conditions and domain regularity.

Specifically, for one-dimensional problems with non-periodic boundaries, the spatial derivatives of each scalar component $u_q$ are evaluated independently. We employ central difference schemes for interior points:
    \begin{gather}
        \frac{\partial u_q(x,t)}{\partial x} \approx \frac{u_q(x+\Delta x, t) - u_q(x-\Delta x, t)}{2\Delta x}, \\
        \frac{\partial^2 u_q(x,t)}{\partial x^2} \approx \frac{u_q(x+\Delta x, t) - 2u_q(x,t) + u_q(x-\Delta x, t)}{(\Delta x)^2}.
    \end{gather}
At domain boundaries, we utilise second-order forward or backward difference stencils to maintain accuracy:
    \begin{gather}
        \frac{\partial u_q(x,t)}{\partial x} \approx \frac{-3u_q(x,t) + 4u_q(x+\Delta x, t) - u_q(x+2\Delta x, t)}{2\Delta x}, \\
        \frac{\partial^2 u_q(x,t)}{\partial x^2} \approx \frac{2u_q(x,t) - 5u_q(x+\Delta x, t) + 4u_q(x+2\Delta x, t) - u_q(x+3\Delta x, t)}{(\Delta x)^2}.
    \end{gather}

For problems with periodic boundary conditions, spectral methods are preferred for their superior accuracy. The Fourier transform of the scalar component $u_q(x,t)$ is defined as:
\begin{equation}
    \hat{u}_q(k,t) = \mathcal{F}[u_q(x,t)] = \int_{-\infty}^{\infty} u_q(x,t) e^{-ikx} \, dx,
\end{equation}
where $k$ is the wave number. In Fourier space, the $n$-th order spatial derivative corresponds to multiplication by $(ik)^n$:
\begin{equation}
    \mathcal{F}\left[\frac{\partial^n u_q(x,t)}{\partial x^n}\right] = (ik)^n \hat{u}_q(k,t).
\end{equation}
The spatial derivatives in physical space are then recovered via the inverse Fourier transform:
\begin{equation}
    \frac{\partial^n u_q(x,t)}{\partial x^n} = \mathcal{F}^{-1}[(ik)^n \hat{u}_q(k,t)].
\end{equation}
These numerical approximations allow for the efficient evaluation of PDE residuals across the spatiotemporal grid during training.

\section{Theoretical interpretation of PDE residual regularisation}
\label{sec:residual-theory}

\subsection{One-step residual-to-error estimate}

The role of the PDE residual can be interpreted through the stability of the discrete one-step problem.
Let $\mathcal{S}_h(\bm{u};\bm{a})$ denote the reference one-step map
and let $\mathcal{R}_{\Delta t,h}(\bm{u},\bm{v};\bm{a})$ denote the discrete PDE residual
when $\bm{u}$ is the current state and $\bm{v}$ is a candidate next state.
Let $\mathcal{U}$ and $\mathcal{Y}$ denote the discrete solution and residual spaces, respectively.
We assume that the discrete one-step residual operator is locally inverse-stable
on a bounded set containing the reference and predicted states,
such that, for admissible states $\bm{v}$ and $\bm{w}$,
\begin{equation}
    \|\bm{v}-\bm{w}\|_{\mathcal{U}}
    \leq
    C_{\mathrm{stab}}
    \left\|
    \mathcal{R}_{\Delta t,h}(\bm{u},\bm{v};\bm{a})
    -\mathcal{R}_{\Delta t,h}(\bm{u},\bm{w};\bm{a})
    \right\|_{\mathcal{Y}}.
    \label{eq:inverse-stability-bound}
\end{equation}
The constant $C_{\mathrm{stab}}$ is assumed to remain bounded in this neighbourhood
and may depend on the states, parameters, discretisation, and chosen norms.
The admissible states share the prescribed boundary conditions;
otherwise, the boundary mismatch contributes an additional error term.
Because the residual and reference solvers may use different discretisations,
$\mathcal{S}_h$ does not necessarily have zero residual.
We therefore define its consistency defect as
\begin{equation}
    \tau_{\Delta t,h}(\bm{u},\bm{a})
    = \left\|
    \mathcal{R}_{\Delta t,h}
    \left(\bm{u},\mathcal{S}_h(\bm{u};\bm{a});\bm{a}\right)
    \right\|_{\mathcal{Y}}.
\end{equation}
Taking $\bm{v}=\mathcal{G}_{\bm{\theta}}(\bm{u};\bm{a})$
and $\bm{w}=\mathcal{S}_h(\bm{u};\bm{a})$ gives
\begin{equation}
    \left\|
    \mathcal{G}_{\bm{\theta}}(\bm{u};\bm{a})
    -\mathcal{S}_h(\bm{u};\bm{a})
    \right\|_{\mathcal{U}}
    \leq C_{\mathrm{stab}}
    \left[
    \left\|
    \mathcal{R}_{\Delta t,h}
    \left(
    \bm{u},\mathcal{G}_{\bm{\theta}}(\bm{u};\bm{a});\bm{a}
    \right)
    \right\|_{\mathcal{Y}}
    +\tau_{\Delta t,h}(\bm{u},\bm{a})
    \right].
    \label{eq:residual-one-step-bound}
\end{equation}
The residual loss therefore controls a contribution to the one-step operator error,
in addition to the direct supervision supplied by the data loss.
Residual-to-error estimates have been established rigorously for restricted classes of linear parabolic PDEs
\cite{deryckErrorAnalysisPhysics2021}.
For the nonlinear and coupled phase-field systems considered here,
Equation~\eqref{eq:residual-one-step-bound} applies locally
under the stated stability and consistency assumptions.

\subsection{Solution-operator continuity and geometric regularisation}

An immediate consequence concerns the continuity of the learned solution operator.
Let $\bm{z}=(\bm{u},\bm{a})$ denote the joint state--parameter input,
use $\mathcal{G}_{\bm{\theta}}(\bm{z})$, $\mathcal{S}_h(\bm{z})$,
and $\tau_{\Delta t,h}(\bm{z})=\tau_{\Delta t,h}(\bm{u},\bm{a})$ as shorthand,
and define
\begin{gather}
    r_{\bm{\theta}}(\bm{z})
    = \left\|
    \mathcal{R}_{\Delta t,h}
    \left(
    \bm{u},\mathcal{G}_{\bm{\theta}}(\bm{u};\bm{a});\bm{a}
    \right)
    \right\|_{\mathcal{Y}}, \\
    \varepsilon_{\bm{\theta}}(\bm{z})
    = C_{\mathrm{stab}}
    \left[
    r_{\bm{\theta}}(\bm{z})
    +\tau_{\Delta t,h}(\bm{z})
    \right].
\end{gather}
Equation~\eqref{eq:residual-one-step-bound} then gives
$\|\mathcal{G}_{\bm{\theta}}(\bm{z})-\mathcal{S}_h(\bm{z})\|_{\mathcal{U}}
\leq\varepsilon_{\bm{\theta}}(\bm{z})$.
Suppose further that the reference solution operator is locally Lipschitz continuous,
\begin{equation}
    \|\mathcal{S}_h(\bm{z}_1)-\mathcal{S}_h(\bm{z}_2)\|_{\mathcal{U}}
    \leq L_{\mathcal{S}}\|\bm{z}_1-\bm{z}_2\|.
\end{equation}
The triangle inequality yields
\begin{equation}
    \|\mathcal{G}_{\bm{\theta}}(\bm{z}_1)
    -\mathcal{G}_{\bm{\theta}}(\bm{z}_2)\|_{\mathcal{U}}
    \leq L_{\mathcal{S}}\|\bm{z}_1-\bm{z}_2\|
    +\varepsilon_{\bm{\theta}}(\bm{z}_1)
    +\varepsilon_{\bm{\theta}}(\bm{z}_2).
    \label{eq:operator-continuity-bound}
\end{equation}
Equation~\eqref{eq:operator-continuity-bound} shows that a uniformly small residual
allows the learned operator to inherit the local continuity of the physical solution operator
up to a residual-dependent error floor.

The residual also provides a geometric regularisation of the learned operator.
For an input domain $\mathcal{Z}$
on which the local inverse-stability estimate holds uniformly,
define the physical and learned solution graphs as
\begin{equation}
    \mathcal{M}_h
    = \left\{(\bm{z},\mathcal{S}_h(\bm{z})):\bm{z}\in\mathcal{Z}\right\},
    \qquad
    \mathcal{M}_{\bm{\theta}}
    = \left\{(\bm{z},\mathcal{G}_{\bm{\theta}}(\bm{z})):\bm{z}\in\mathcal{Z}\right\}.
\end{equation}
At a fixed input $\bm{z}$, their vertical distance satisfies
\begin{equation}
    d_{\bm{z}}(\mathcal{M}_{\bm{\theta}},\mathcal{M}_h)
    := \|\mathcal{G}_{\bm{\theta}}(\bm{z})-\mathcal{S}_h(\bm{z})\|_{\mathcal{U}}
    \leq C_{\mathrm{stab}}
    \left[r_{\bm{\theta}}(\bm{z})+\tau_{\Delta t,h}(\bm{z})\right].
    \label{eq:manifold-pointwise-bound}
\end{equation}
Equation~\eqref{eq:manifold-pointwise-bound} shows that the PDE residual
evaluated at an input provides an upper bound on the prediction error
at that input without requiring its labelled output.
If $r_{\bm{\theta}}(\bm{z})\leq\epsilon$
and $\tau_{\Delta t,h}(\bm{z})\leq\bar\tau$
uniformly on $\mathcal{Z}$, then
\begin{equation}
    d_{\mathrm{H}}(\mathcal{M}_{\bm{\theta}},\mathcal{M}_h)
    \leq C_{\mathrm{stab}}(\epsilon+\bar\tau),
    \label{eq:manifold-hausdorff-bound}
\end{equation}
where the product-space metric uses the solution norm $\|\cdot\|_{\mathcal{U}}$
for points with the same input.
An averaged version follows for an input distribution $\rho$.
Defining
\begin{equation}
    \mathcal{L}_{\mathrm{p},\rho}
    = \mathbb{E}_{\bm{z}\sim\rho}\left[r_{\bm{\theta}}(\bm{z})^2\right],
    \qquad
    \mathcal{T}_{\rho}
    = \mathbb{E}_{\bm{z}\sim\rho}\left[\tau_{\Delta t,h}(\bm{z})^2\right],
\end{equation}
Equation~\eqref{eq:manifold-pointwise-bound} gives
\begin{equation}
    \mathbb{E}_{\bm{z}\sim\rho}
    \left[
    \|\mathcal{G}_{\bm{\theta}}(\bm{z})-\mathcal{S}_h(\bm{z})\|_{\mathcal{U}}^2
    \right]
    \leq 2C_{\mathrm{stab}}^2
    \left(\mathcal{L}_{\mathrm{p},\rho}+\mathcal{T}_{\rho}\right).
    \label{eq:manifold-mean-bound}
\end{equation}
The data loss samples the graph distance directly at labelled inputs,
while the residual supplies the additional physics-based control quantified above.
Thus, minimising the residual reduces an upper bound on the distance
between the learned and physical solution graphs over the inputs where the residual is controlled
\cite{wangLearningSolutionOperator2021,liPhysicsInformedNeuralOperator2023}.

\subsection{Autoregressive error propagation}

The connection with long-term prediction follows from the propagation of one-step defects.
Let the reference and predicted trajectories satisfy
$\bm{u}_{n+1}=\mathcal{S}_h(\bm{u}_n;\bm{a})$
and $\hat{\bm{u}}_{n+1}=\mathcal{G}_{\bm{\theta}}(\hat{\bm{u}}_n;\bm{a})$,
respectively, and define $\bm{e}_n=\hat{\bm{u}}_n-\bm{u}_n$.
If $\mathcal{S}_h$ is locally Lipschitz with constant $q$ and
\begin{equation}
    \delta_n = \left\|
    \mathcal{G}_{\bm{\theta}}(\hat{\bm{u}}_n;\bm{a})
    -\mathcal{S}_h(\hat{\bm{u}}_n;\bm{a})
    \right\|_{\mathcal{U}}
\end{equation}
is the local learned-update defect, then
\begin{equation}
    \|\bm{e}_{n+1}\|_{\mathcal{U}}
    \leq q\|\bm{e}_n\|_{\mathcal{U}}+\delta_n,
\end{equation}
and recursive application yields
\begin{equation}
    \|\bm{e}_N\|_{\mathcal{U}}
    \leq q^N\|\bm{e}_0\|_{\mathcal{U}}
    +\sum_{j=0}^{N-1}q^{N-1-j}\delta_j.
    \label{eq:autoregressive-error-bound}
\end{equation}
Equation~\eqref{eq:autoregressive-error-bound} shows that reducing
the local one-step defect $\delta_j$ lowers its contribution to the long-horizon error.
For cases using physics-informed one-step training,
the learned update has smaller one-step errors on the reference-state distribution,
and this improvement generalises to the nearby predicted states encountered during rollout.
For cases using rollout fine-tuning,
the residual is also evaluated on model-generated states.
To quantify this mechanism, define
\begin{gather}
    r_j = \left\|
    \mathcal{R}_{\Delta t,h}
    \left(
    \hat{\bm{u}}_j,
    \mathcal{G}_{\bm{\theta}}(\hat{\bm{u}}_j;\bm{a});
    \bm{a}
    \right)
    \right\|_{\mathcal{Y}},
    \qquad
    \tau_j = \tau_{\Delta t,h}(\hat{\bm{u}}_j,\bm{a}), \\
    \mathcal{L}_{\mathrm{p}}^{\mathrm{roll}}
    = \frac{1}{N}\sum_{j=0}^{N-1}r_j^2,
    \qquad
    \mathcal{T}^{\mathrm{roll}}
    = \frac{1}{N}\sum_{j=0}^{N-1}\tau_j^2.
\end{gather}
Equation~\eqref{eq:residual-one-step-bound} gives
$\delta_j\leq C_{\mathrm{stab}}(r_j+\tau_j)$.
Substitution into Equation~\eqref{eq:autoregressive-error-bound}
and application of the Cauchy--Schwarz inequality yield
\begin{equation}
    \|\bm{e}_N\|_{\mathcal{U}}
    \leq q^N\|\bm{e}_0\|_{\mathcal{U}}
    + C_{\mathrm{stab}}\sqrt{N}\,\Gamma_N(q)
    \left[
    \sqrt{\mathcal{L}_{\mathrm{p}}^{\mathrm{roll}}}
    +\sqrt{\mathcal{T}^{\mathrm{roll}}}
    \right],
    \label{eq:rollout-residual-bound}
\end{equation}
where
\begin{equation}
    \Gamma_N(q)
    = \left(\sum_{k=0}^{N-1}q^{2k}\right)^{1/2}
    = \begin{cases}
        \displaystyle
        \left(\dfrac{1-q^{2N}}{1-q^2}\right)^{1/2}, & q\neq 1, \\[2mm]
        \sqrt{N}, & q=1.
    \end{cases}
\end{equation}
For fixed $C_{\mathrm{stab}}$, $q$, and consistency defect,
Equation~\eqref{eq:rollout-residual-bound} makes the contribution of the physics loss explicit:
the upper bound associated with learned-update defects scales with
$\sqrt{\mathcal{L}_{\mathrm{p}}^{\mathrm{roll}}}$.
Rollout fine-tuning evaluates and minimises
$\mathcal{L}_{\mathrm{p}}^{\mathrm{roll}}$ directly.
The amplification factor $q$ remains governed by the reference evolution map,
so global contractivity or energy stability requires separate control
\cite{mccabeStabilityAutoregressiveNeural2023}.
This distinction is important for spinodal decomposition,
where the amplification of some perturbation modes is part of the physical dynamics rather than a numerical instability.

\section{Performance evaluation metrics}
\label{sec:metrics}

Model accuracy is quantified using two complementary metrics:
the relative $L^2$ error and the relative Hausdorff distance.
The relative $L^2$ error quantifies the scale-normalised global discrepancy
over the full spatio-temporal field
and captures error accumulation during autoregressive prediction.
Because the relative $L^2$ error aggregates discrepancies over the full
spatio-temporal domain, localised errors near the diffuse interface may be
underrepresented in this global measure.
To account for this limitation,
the relative Hausdorff distance complements the global measure
by quantifying the maximum geometric discrepancy
between the predicted and reference interfaces
\cite{huttenlocherComparingImagesHausdorff1993,
tahaMetricsEvaluating3D2015}.
Normalisation by the mesh size expresses this discrepancy
in units of grid resolution
and enables comparison across the benchmark problems.
Together, the two metrics evaluate both field-value fidelity
and interface-location accuracy.

The relative $L^2$ error measures the global discrepancy between predicted and reference fields over the entire spatio-temporal domain. It is defined as:
\begin{equation}
    \text{Rel. }L^2 = \frac{1}{N_{\text{test}}} \frac{1}{N_{\text{chnl}}} \sum_{i=1}^{N_{\text{test}}} \sum_{j=1}^{N_{\text{chnl}}} \frac{
        \| \hat{u}_{i,j} - u_{i,j} \|_2
    }{
        \| u_{i,j} \|_2
    }, \label{eq:rel_L2}
\end{equation}
where $N_{\text{test}}$ and $N_{\text{chnl}}$ denote the number of test cases and variable channels, respectively, and the scalar quantities $\hat{u}_{i,j}$ and $u_{i,j}$ correspond to the predicted and reference trajectories for channel $j$ of test case $i$.

To assess the geometric accuracy of the predicted microstructures, we employ
the relative Hausdorff distance. Let $\mathcal{I}_{\text{pred}}$ and
$\mathcal{I}_{\text{ref}}$ denote the sets of points constituting the predicted
and reference interfaces ($\phi=0.5$ for corrosion, $\phi=0$ for
solidification, and $c=0$ for spinodal decomposition), respectively. The
Hausdorff distance $d_{\text{H}}$ is defined as:
\begin{equation}
    d_{\text{H}}(\mathcal{I}_{\text{pred}}, \mathcal{I}_{\text{ref}}) = \max \left\{ \sup_{p \in \mathcal{I}_{\text{pred}}} \inf_{r \in \mathcal{I}_{\text{ref}}} \|p - r\|_2, \, \sup_{r \in \mathcal{I}_{\text{ref}}} \inf_{p \in \mathcal{I}_{\text{pred}}} \|r - p\|_2 \right\}.
\end{equation}
The relative Hausdorff distance is then computed by normalising $d_{\text{H}}$ with the computational mesh size $\Delta x$ and averaged over the test set:
\begin{equation}
    \text{Rel. } d_{\text{H}} = \frac{1}{N_{\text{test}}}\sum_{i=1}^{N_{\text{test}}}  
    \frac{d_{i, \text{H}}(\mathcal{I}_{\text{pred}}, \mathcal{I}_{\text{ref}})}{\Delta x}.
\end{equation}
This dimensionless metric quantifies the maximum interface positioning error in units of grid spacing.

\section{Benchmark descriptions and governing equations}
\label{sec:benchmark-description}

\subsection{Pencil-electrode and electro-polishing corrosion}

The first two benchmarks focus on electrochemical corrosion processes, which are classical applications of phase-field modelling in materials science. For the 1D pencil-electrode corrosion, we parametrise the problem by varying the interface kinetics coefficient $L$ to control the corrosion rate and evaluate the model's ability to generalise across different kinetic regimes. For the 2D electro-polishing corrosion benchmark, the parametric space is defined by the randomly sampled initial interface profiles, which directly influence the surface roughness evolution during corrosion. 

The corrosion process is modelled using the Kim-Kim-Suzuki (KKS) phase-field formulation \cite{kimPhasefieldModelBinary1999}. A continuous phase-field variable $\phi$ differentiates the solid metal phase ($\phi=1$) from the liquid electrolyte ($\phi=0$). The evolution of $\phi$ follows an Allen-Cahn type equation:
\begin{equation}
    \frac{\partial \phi(\bm{x}, t)}{\partial t}=-L \frac{\delta \mathcal{E}}{\delta \phi},\label{eq:AC}
\end{equation}%
coupled with a Cahn-Hilliard equation describing the evolution of the concentration field $c$:
\begin{equation}
    \frac{\partial c (\bm{x}, t) }{\partial t}  =\nabla \cdot M \nabla \frac{\delta \mathcal{E}}{\delta c},\label{eq:CH}
\end{equation}%
where $L$ and $M$ denote the interface kinetics coefficient and diffusivity, respectively. The total free energy functional $\mathcal{E}$ is defined as \cite{maiPhaseFieldModel2016}:
\begin{equation}
    \mathcal{E}(\phi, c) = \int_{\Omega} \left[
        f(\phi, c) + \frac{\alpha_\phi}{2} \lvert \nabla \phi \rvert ^2
        \right] \, \mathrm{d}\bm{x},
    \label{eq:total_free_energy}
\end{equation}
with the local free energy density $f(\phi, c)$ given by:
\begin{equation}
    f(\phi, c) = \mathcal{A}\left[c-h(\phi)\left(c_{\mathrm{Se}}-c_{\mathrm{Le}}\right)-c_{\mathrm{Le}}\right]^{2}+w_\phi g(\phi),
    \label{eq:free_energy_density}
\end{equation}
where $\mathcal{A}$ parametrises the free energy density, and
$c_{\mathrm{Se}}$ and $c_{\mathrm{Le}}$ represent the normalised equilibrium
concentrations for the solid and liquid phases, respectively. The double-well
potential and interpolation functions are given by
$g(\phi)=\phi^{2}(1-\phi)^{2}$ and $h(\phi)=-2\phi^{3}+3\phi^{2}$. The
parameters $w_\phi$ and $\alpha_\phi$ determine the potential barrier height
and the gradient energy coefficient. Detailed physical parameters are listed
in Table~\ref{tab:corrosion-parameters}.

A simplified form of the governing PDEs can be derived by explicitly computing the variational derivatives $\delta \mathcal{E}/\delta \phi$ and $\delta \mathcal{E}/\delta c$:
    \begin{align}
        \text{Cahn--Hilliard: } & \frac{\partial c}{\partial t}- 2\mathcal{A}M \Delta c + 2\mathcal{A}M \left(c_{\mathrm{Se}}-c_{\mathrm{Le}}\right) \Delta h\left(\phi\right) =0 , \label{eq:chcorro} \\
        \text{Allen--Cahn: }    & \frac{\partial \phi}{\partial t}
        -2\mathcal{A}L\left[c-h(\phi)\left(c_{\mathrm{Se}}-c_{\mathrm{Le}}\right)-c_{\mathrm{Le}}\right]\left(c_{\mathrm{Se}}-c_{\mathrm{Le}}\right) h^{\prime}(\phi) +L w_\phi g^{\prime}(\phi) - L \alpha_\phi \Delta \phi =0.
        \label{eq:accorro}
    \end{align}

\subsection{Dendritic crystal solidification}
\label{sec:solidification-description}

Dendrites are intricate tree-like microstructures that commonly form in frozen alloys and supercooled liquids during solidification \cite{karmaPhasefieldModelDendritic1999}. The dendritic growth process is governed by the coupled evolution of the phase-field variable $\phi$ and the temperature field $T$, described by the following equations \cite{karmaPhasefieldModelDendritic1999,yangEfficientLinearStabilized2019}:
    \begin{gather}
        \rho(\phi)\frac{\partial \phi}{\partial t} = -\frac{\delta \mathcal{E}}{\delta \phi} - \frac{\lambda}{\varepsilon} h'(\phi) T, \\
        \frac{\partial T}{\partial t} = \nabla\cdot \left(D\nabla T\right) + K h'(\phi) \frac{\partial \phi}{\partial t},
    \end{gather}
where $\phi$ distinguishes between the solid ($\phi=1$) and liquid ($\phi=-1$) phases, and $T$ denotes the normalised temperature field. The mobility parameter $\rho(\phi) > 0$ is assumed constant in this work. Furthermore, $\varepsilon$ denotes the interface width parameter, $\lambda$ is the linear kinetic coefficient, $D$ is the thermal diffusivity, and $K$ is the latent heat coefficient governing the rate of heat release during solidification. The interpolation function $h(\phi)$, related to latent heat generation, is defined as:
\begin{equation}
    h(\phi) = \frac{1}{5}\phi^5 - \frac{2}{3}\phi^3 + \phi.
\end{equation}

The system free energy functional $\mathcal{E}$ is given by \cite{karmaQuantitativePhasefieldModeling1998}:
\begin{equation}
    \mathcal{E}\left(\phi, T\right) = \int_{\Omega} \left[
        \frac{1}{2} \kappa^2 \left(\nabla\phi\right) \lvert  \nabla\phi\rvert ^2
        + \frac{1}{\varepsilon^2} F(\phi)
        + \frac{\lambda}{2\varepsilon K} T^2
        \right] \, \mathrm{d}\bm{x},
\end{equation}
where $F(\phi) = \frac{1}{4}(1-\phi^2)^2$ is the double-well Ginzburg-Landau potential. The anisotropy in interface energy is introduced via the coefficient $\kappa(\theta)$:
\begin{equation}
    \kappa(\theta) = 1 + \sigma \cos(m\theta),
\end{equation}
where $\sigma$ is the anisotropy strength parameter, $m$ is the anisotropy mode, and $\theta=\arctan\left(\phi_y / \phi_x\right)$ represents the angle between the interface normal and the x-axis. The variational derivative $\frac{\delta \mathcal{E}}{\delta \phi}$ is derived as:
\begin{equation}
    \frac{\delta \mathcal{E}}{\delta \phi} = -\nabla \cdot \left[
        \kappa^2(\theta) \nabla \phi
        + \kappa(\theta) \lvert \nabla \phi \rvert ^2
        \bm{H}(\phi)
        \right] + \frac{f(\phi)}{\varepsilon^2},
\end{equation}
where $f(\phi) = F'(\phi)$. The vector term $\bm{H}(\phi)$ arises from the anisotropy. Setting the anisotropy mode to $m=4$ generates four-fold symmetric dendritic patterns, with $\bm{H}(\phi)$ given by:
\begin{equation}
    \bm{H}(\phi) = \frac{16\sigma}{\lvert \nabla \phi \rvert ^6}
    \left(
    \phi_x^3 \phi_y ^2 - \phi_x \phi_y^4,
    \phi_y^3 \phi_x ^2 - \phi_y \phi_x^4
    \right).
\end{equation}


\subsection{Spinodal decomposition}

Spinodal decomposition is a fundamental phase separation mechanism in binary alloys and polymer blends, whereby a homogeneous mixture spontaneously separates into distinct coexisting phases \cite{krekhovPhaseSeparationPresence2004}. The evolution of the concentration field $c$ is governed by the Cahn-Hilliard equation:
\begin{equation}
    \frac{\partial c}{\partial t} = \nabla \cdot\left(
    M\nabla \mu
    \right),
\end{equation}
where $M$ denotes the mobility parameter and $\mu$ is the chemical potential derived from the total free energy functional:
\begin{equation}
    \mathcal{E}(c) = \int_{\Omega} \left[
        f(c) + \frac{\lambda}{2} \lvert \nabla c \rvert ^2
        \right] \, \mathrm{d}\bm{x},
\end{equation}
with $f(c) = \frac{1}{4}(c^2-1)^2$ being the double-well potential. The chemical potential $\mu$ is defined as the variational derivative:
\begin{equation}
    \mu = \frac{\delta \mathcal{E}}{\delta c} = f'(c) - \lambda \nabla^2 c.
\end{equation}
Periodic boundary conditions are applied in both spatial directions.


\section{Numerical implementation and dataset generation}
\label{sec:numerical-implementation-dataset-generation}

High-fidelity reference datasets are generated using numerical solvers tailored to each benchmark problem. Specifically, we employ FEM via the FEniCS library for the corrosion and solidification examples, while a pseudo-spectral method is utilised for the spinodal decomposition simulation. For each problem, we conducted parametric sweeps across a diverse range of physical parameters and initial conditions to produce multiple spatiotemporal solution trajectories.

To construct the training data, these trajectories are processed into one-step input-output pairs mapping the system state from time $t_n$ to $t_{n+1}$. The input tensors are formed by concatenating the instantaneous solution fields (e.g., phase-field variable $\phi$, concentration $c$, temperature $T$) with spatially invariant parameter channels and coordinate grids, thereby explicitly embedding the physical configuration and domain geometry. The aggregated dataset is randomly partitioned into training and validation subsets with a 75/25 ratio. To rigorously assess generalisation capability, we reserved distinct test sets containing parameter values and initial configurations strictly excluded from the training distribution, which are used to evaluate the models' long-term stability via autoregressive prediction.

\subsection{Pencil-electrode corrosion}
\label{sec:pencil-electrode-corrosion-implementation}

The weak form of the governing Allen-Cahn and Cahn-Hilliard equations (Eqs.~\eqref{eq:AC} and \eqref{eq:CH}) with backward Euler time discretisation and test functions $v_\phi$ and $v_c$ is given by:
\begin{equation}
    \int_{\Omega} \frac{c^{n+1}-c^{n}}{\Delta t} v_c \,\mathrm{d}\Omega
    + 2\mathcal{A}M \int_{\Omega} \nabla c^{n+1} \cdot \nabla v_c \,\mathrm{d}\Omega
    - 2\mathcal{A}M \int_{\Omega} \left(c_{\mathrm{Se}}-c_{\mathrm{Le}}\right) \nabla h\left(\phi^{n+1}\right) \cdot \nabla v_c \,\mathrm{d}\Omega = 0,
    \label{eq:cahn-hilliard-weak-form}
\end{equation}
and
\begin{equation}
    \begin{aligned}
        \int_{\Omega} \frac{\phi^{n+1}-\phi^{n}}{\Delta t} v_\phi \,\mathrm{d}\Omega
        - 2\mathcal{A}L \int_{\Omega} \left[c^{n+1}-h(\phi^{n+1})\left(c_{\mathrm{Se}}-c_{\mathrm{Le}}\right)-c_{\mathrm{Le}}\right]\left(c_{\mathrm{Se}}-c_{\mathrm{Le}}\right) h^{\prime}(\phi^{n+1}) v_\phi \,\mathrm{d}\Omega
                                                                                               &      \\
        + L w_\phi \int_{\Omega} g^{\prime}(\phi^{n+1}) v_\phi \,\mathrm{d}\Omega
        + L \alpha_\phi \int_{\Omega} \nabla \phi^{n+1} \cdot \nabla v_\phi \,\mathrm{d}\Omega & = 0.
        \label{eq:allen-cahn-weak-form}
    \end{aligned}
\end{equation}
In FEniCS, we directly sum Equation~\eqref{eq:cahn-hilliard-weak-form} and Equation~\eqref{eq:allen-cahn-weak-form} to form a coupled nonlinear variational problem, which is solved using the Newton-Raphson method at each time step.

The simulation domain is one-dimensional with length \qty{100}{\mu\meter}, centred at the pencil electrode. The initial metal-electrolyte interface is positioned at $x=\qty{35}{\mu\meter}$. We discretise the spatial domain using linear Lagrange elements on a uniform mesh with 100 elements. The time step is set to \qty{1}{s}, and simulations are run for a total duration of \qty{100}{s}. The training and validation dataset comprises only 5 different values of the interface kinetics coefficient $L$ of \numlist{1.0e-9;1.0e-8;1.0e-7;1.0e-5;1.0e0} \unit{\cubic\meter/(\joule\second)}, resulting in 500 one-step input-output pairs. For testing, we select 6 unseen values of $L$ of \numlist{5.0e-9;2.5e-8;5.0e-7;1.0e-6;1.0e-3;5.0e-1} \unit{\cubic\meter/(\joule\second)}.

\subsection{Electro-polishing corrosion}

The weak form of the governing equations (Eqs.~\eqref{eq:AC} and \eqref{eq:CH}) with backward Euler time discretisation and test functions $v_\phi$ and $v_c$ is identical to that in Section~\ref{sec:pencil-electrode-corrosion-implementation}. The spatial domain is two-dimensional with \qty{100}{\mu\meter} for x-direction and \qty{50}{\mu\meter} for y-direction. We discretise the domain using linear Lagrange elements on a uniform mesh with \numproduct{100 x 50} elements. The total simulation time is \qty{2e4}{s} with a time step of \qty{200}{s}.

The training and validation dataset consists of 10 different initial interface morphologies generated by superimposing \num{10} sinusoidal perturbation modes with a fixed interface kinetics coefficient $L=\qty{1e-10}{\cubic\meter/(\joule\second)}$. The random amplitudes $a_k$ are sampled from a Gaussian distribution $\mathcal{N}(0, \sigma_a/\sqrt{k})$ with $\sigma_a=\qty{2.0}{\mu\meter}$. The dataset contains \num{1000} one-step input-output pairs. For testing, we generate 5 unseen initial interface profiles using the same procedure with different random seeds.

\subsection{Dendritic crystal solidification}
\label{sec:dendritic-solidification-implementation}

The governing equations are solved using a linearised semi-implicit time
integration scheme to decouple the evolution of $\phi$ and $T$. At each time
step, we first update the phase-field $\phi^{n+1}$ via a linear variational
problem where the nonlinear source term
$s(\phi)=-f(\phi)=\phi-\phi^3$ is linearised around $\phi^n$ using a
first-order Taylor expansion. Specifically,
\begin{equation}
    s(\phi^{n+1}) \approx s(\phi^n) + s'(\phi^n)(\phi^{n+1} - \phi^n)
    = (1 - 3(\phi^n)^2)\phi^{n+1} + 2(\phi^n)^3.
\end{equation}
The anisotropy terms and temperature coupling are treated explicitly. The weak form for the phase-field update is given by:
\begin{equation}
    \begin{aligned}
        \int_{\Omega} \rho \frac{\phi^{n+1}-\phi^{n}}{\Delta t} v_\phi \,\mathrm{d}\Omega
        + \int_{\Omega} \kappa^2(\theta^n) \nabla \phi^{n+1} \cdot \nabla v_\phi \,\mathrm{d}\Omega
        + \int_{\Omega} \kappa(\theta^n) \lvert \nabla \phi^n \rvert ^2 \bm{H}(\phi^n) \cdot \nabla v_\phi \,\mathrm{d}\Omega &      \\
        - \frac{1}{\varepsilon^2} \int_{\Omega} \left[ (1 - 3(\phi^n)^2)\phi^{n+1} + 2(\phi^n)^3 \right] v_\phi \,\mathrm{d}\Omega
        + \frac{\lambda}{\varepsilon} \int_{\Omega} h'(\phi^n) T^n v_\phi \,\mathrm{d}\Omega                                      & = 0,
    \end{aligned}
\end{equation}
where $v_\phi$ is the test function. Subsequently, the temperature field $T^{n+1}$ is updated by solving the heat equation implicitly, utilising the newly computed $\phi^{n+1}$ for the latent heat source term. The corresponding weak form is:
\begin{equation}
    \int_{\Omega} \frac{T^{n+1}-T^{n}}{\Delta t} v_T \,\mathrm{d}\Omega
    + D \int_{\Omega} \nabla T^{n+1} \cdot \nabla v_T \,\mathrm{d}\Omega
    - K \int_{\Omega} h'(\phi^{n+1}) \frac{\phi^{n+1}-\phi^{n}}{\Delta t} v_T \,\mathrm{d}\Omega = 0,
\end{equation}
where $v_T$ is the test function. Both linear systems are solved using the GMRES method with an ILU preconditioner. This staggered approach effectively decouples the two fields while maintaining numerical stability.


The simulation is performed on a two-dimensional square domain $\Omega = [-1,1]\times[-1,1]$ with homogeneous Neumann boundary conditions. The spatial domain is discretised using linear Lagrange elements on a uniform crossed mesh with \numproduct{128 x 128} subdivisions. In this semi-implicit scheme, we set the time step to $\Delta t = 0.01 \unit{s}$. The total simulation time is $T=10 \unit{s}$. Since the fully implicit scheme is adopted in PINOs, we loosen the time step restriction to $\Delta t = 0.05 \unit{s}$. The training and validation dataset is generated using 5 latent heat coefficients $K \in \{0.8, 1.0, 1.2, 1.4, 1.6\}$, leading to \num{1000} one-step input-output pairs. For testing, we select 4 unseen values $K \in \{0.9, 1.3, 1.7, 2.0\}$. It is worth noting that to evaluate the extrapolation capability of the PF-PINO model, we extend the testing range of $K$ beyond that of the training set.

\subsection{Spinodal decomposition}

Unlike the previous benchmarks, the spinodal decomposition problem is solved using a pseudo-spectral method with periodic boundary conditions. The concentration field is discretised on a uniform grid of \numproduct{64 x 64} points. Time integration is performed using a Crank-Nicolson scheme for enhanced accuracy and stability. The discrete form of the Cahn-Hilliard equation in Fourier space is given by:
\begin{equation}
    \frac{c^{n+1} - c^n}{\Delta t} = -M k^2 \left[
        \frac{1}{2} \left( \hat{\mu}^{n+1} + \hat{\mu}^n \right)
        \right] = -M k^2 \left[
        \frac{1}{2} \left( \widehat{f'(c^{n+1})} + \widehat{f'(c^n)} \right) + \lambda k^2 \frac{c^{n+1} + c^n}{2}
        \right],
\end{equation}
where $k$ denotes the wave number in Fourier space, and $\hat{\mu}$ is the Fourier transform of the chemical potential $\mu$.

To resolve the nonlinearity arising from $f'(c^{n+1})$, a Picard fixed-point iteration is employed. Given an initial guess $c^{n+1,(0)} = c^n$, the iterative update is formulated as:
\begin{equation}
    c^{n+1,(m+1)} = \frac{
        (2-\Delta t M \lambda k^4) c^n - \Delta t M k^2\left[
            \widehat{f'(c^{n+1,(m)})} + \widehat{f'(c^n)}
            \right]
    }{
        2 + \Delta t M \lambda k^4
    },\quad (m=0,1,2,\ldots)
\end{equation}
where $(m)$ is the iteration index. The iteration continues until convergence is achieved, defined by $\lVert c^{n+1,(m+1)} - c^{n+1,(m)} \rVert < \epsilon$, with a tolerance $\epsilon = \num{1e-9}$.

\begin{figure}[htbp]
    \centering
    \includegraphics[width=0.3\textwidth]{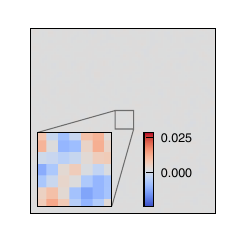}
    \caption{An example of the initial concentration field for the spinodal decomposition benchmark, generated by band-limited Fourier modes.}
    \label{fig:spinodal-initial-condition}
\end{figure}

The spatial domain $\Omega=[0,1]^2$ is discretised using a uniform grid with \numproduct{64 x 64} points. The time step is set to $\Delta t = \qty{5e-5}{s}$ , and the total simulation time is \qty{5e-3}{s}. The parametrised mobility $M$ is randomly sampled from a uniform distribution $U(0.5, 1.5)$. The initial concentration perturbations are generated using \num{100} Fourier modes with wave numbers bounded between $k_{\text{min}}=-15$ and $k_{\text{max}}=15$. The amplitudes are set to be $a_i = 0.01/N_{\text{pert}}=\num{1e-4}$ to ensure small initial fluctuations. The perturbation phases $\phi_i$ are uniformly sampled from $[0, 2\pi]$ (shown in Figure~\ref{fig:spinodal-initial-condition}). The mean concentration is set to $c_0=0$ to induce a symmetric phase separation scenario. We generate \num{20} different initial concentration fields and mobility values to create a training and validation dataset of \num{2000} one-step input-output pairs. For testing, we generate \num{5} unseen initial concentration profiles and the mobility values are set to $M \in \{0.6,0.8,1.0,1.2,1.4\}$.

\section{Physical parameters}
\label{sec:physical-parameters}

Physical parameters used in the phase-field models for the corrosion and solidification benchmarks are summarised in Tables~\ref{tab:corrosion-parameters} and \ref{tab:solidification-parameters}.

\begin{table}[htbp]
    \centering
    \caption{Physical parameters used in the phase-field corrosion modelling (SI units). In 1D pencil-electrode corrosion test, the interface kinetic coefficient $L$ is parametrised over a wide range to capture different corrosion regimes, while in 2D electro-polishing corrosion test, $L$ is fixed at \qty{1e-10}{\cubic\meter/(\joule\second)}}
    \label{tab:corrosion-parameters}
    \begin{tblr}{width=\textwidth,colspec={X[1,c]X[5,c]X[1,c]},row{1}={font=\bfseries}}
        \hline
        Notation        & Description                                               & Value                                                                                        \\
        \hline
        $L$             & Interface kinetic coefficient                             & Parametrised                                                                                 \\
        $\alpha_\phi$   & Gradient energy coefficient                               & \num{1.02e-4}                                                                                \\
        $w_\phi$        & Height of the double well potential                       & \num{1.76e7}                                                                                 \\
        $\ell$          & Interface thickness                                       & \num{1.0e-5}                                                                                 \\
        $M$             & Diffusivity parameter                                     & \num{7.94e-18}                                                                               \\
        $\mathcal{A}$   & Free energy density-related parameter                     & \num{5.35e7}                                                                                 \\
        $c_\mathrm{Se}$ & Normalised equilibrium concentration for the solid phase  & \num{1.0}                                                                                    \\
        $c_\mathrm{Le}$ & Normalised equilibrium concentration for the liquid phase & \num{0.036}                                                                                  \\
        \hline
        \SetCell[c=3,r=1]{l}
    \end{tblr}
\end{table}

\begin{table}[htbp]
    \centering
    \caption{Physical parameters used in the phase-field solidification modelling (Non-dimensional units).}
    \label{tab:solidification-parameters}
    \begin{tblr}{width=\textwidth,colspec={X[1,c]X[5,c]X[1,c]},row{1}={font=\bfseries}}
        \hline
        Notation      & Description                   & Value        \\
        \hline
        $K$           & Latent heat coefficient       & Parametrised \\
        $\rho$        & Mobility parameter            & \num{1.0e3}  \\
        $\varepsilon$ & Interface width parameter     & \num{0.015}  \\
        $\lambda$     & Linear kinetic coefficient    & \num{4.0e2}  \\
        $D$           & Temperature diffusivity       & \num{2.5e-3} \\
        $\sigma$      & Anisotropy strength parameter & \num{0.1}    \\
        $m$           & Anisotropy mode               & 4            \\
        \hline
    \end{tblr}
\end{table}

\section{Hyperparameters and training details}
\label{sec:hyperparameters}

Hyperparameter configurations for all benchmark tests are summarised in Table~\ref{tab:hyperparameters-all}. To ensure a controlled comparison, PF-PINO and FNO share identical network architectures, training data, and common training hyperparameters for each benchmark. PF-PINO additionally incorporates PDE residual losses together with gradient normalisation and, for coupled multi-field benchmarks, staggered residual training. All models are implemented using the \texttt{JAX} library with \texttt{Equinox} for neural network construction and \texttt{Optax} for optimisation.
Model training and inference are
conducted on a single NVIDIA GeForce RTX 5090 GPU with \num{32} GB memory.
FEM simulations are conducted on a CPU node with \num{25} vCPUs and an Intel
Xeon Platinum 8470Q processor.

For all benchmarks, we employ the Adam optimiser with an initial learning rate as specified in Table~\ref{tab:hyperparameters-all}. The learning rate is decayed exponentially by a factor indicated by the LR decay rate every LR decay step epochs until it reaches the minimum value. The out-channels of the network correspond to the number of solution fields being predicted, while the in-channels include the solution fields at the current time step, parameter channels, and coordinate grids. The PDE residuals are computed using either finite difference (FD) or spectral methods, as specified in Table~\ref{tab:hyperparameters-all}. 
\begin{table}[htbp]
    \centering
    \caption{Hyperparameter configurations for all benchmark tests.}
    \label{tab:hyperparameters-all}
    \begin{tblr}{
        width=\textwidth,
        colspec={X[0.8,c]X[1.5,c]X[1,c]X[1,c]X[1,c]X[1,c]},
        row{1}={m,font=\bfseries},
        }
        \hline
        Group & Hyperparameter & Corrosion1d & Corrosion2d & Solidification & Spinodal Decomposition \\
        \hline
        \SetCell[r=6]{c} \textbf{Network}
              & In-channels    & 5           & 5           & 5              & 4                      \\
              & Out-channels   & 2           & 2           & 2              & 1                      \\
              & Depth          & 4           & 4           & 4              & 3                      \\
              & Width          & 64          & 64          & 64             & 64                     \\
              & Modes          & 8           & (8,8)       & (16,16)        & (32,32)                \\
              & Activation     & GeLU        & ReLU        & ReLU           & GeLU                   \\
        \hline
        \SetCell[r=8]{c} \textbf{Training}
              & Epochs         & \num{10000} & \num{5000}  & \num{3000}     & \num{3000}             \\
              & Batch size     & 64          & 128         & 64             & 64                     \\
              & Optimiser      & Adam        & Adam        & Adam           & Adam                   \\
              & Initial LR     & \num{1e-3}  & \num{5e-4}  & \num{5e-4}     & \num{5e-4}             \\
              & LR decay rate  & 0.95        & 0.95        & 0.95           & 0.95                   \\
              & LR decay step  & 500         & 200         & 50             & 200                    \\
              & LR min value   & \num{1e-5}  & \num{1e-5}  & \num{1e-5}     & \num{1e-5}             \\
              & PDE residual   & FD          & FD          & FD             & Spectral               \\
        \hline
    \end{tblr}
\end{table}

\section{Additional visualisations}
\label{sec:additional-visualisations}

In this section, we provide additional figures to supplement the results presented in the main text.

\subsection{Pencil-electrode corrosion}
\label{sec:pencil-electrode-corrosion-additional-visualisations}

\begin{figure}[htbp]
    \centering
    \includegraphics{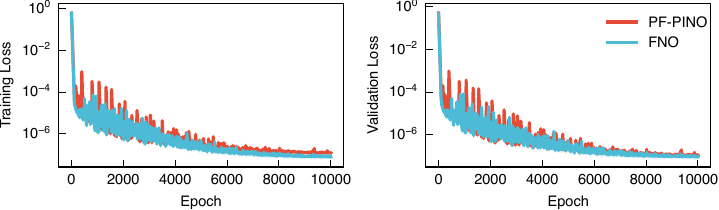}
    \caption{Pencil-electrode corrosion: One-step training and validation loss curves during PF-PINO and FNO training. }
    \label{fig:corrosion1d_training_validation_loss}
\end{figure}

\begin{figure}[htbp]
    \centering
    \includegraphics{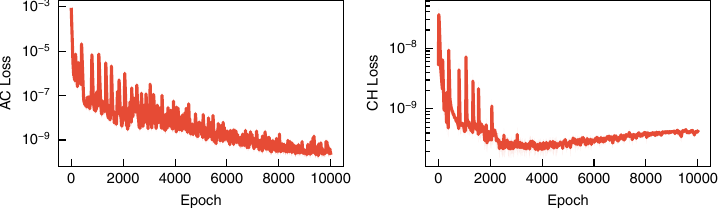}
    \caption{Pencil-electrode corrosion: One-step training loss curves of the Allen-Cahn and Cahn-Hilliard equations during PF-PINO training. }
    \label{fig:corrosion1d_ac_ch_loss}
\end{figure}

Figure~\ref{fig:corrosion1d_training_validation_loss} illustrates the one-step training and validation loss curves during PF-PINO and FNO training for the pencil-electrode corrosion benchmark. Both models exhibit stable convergence in \num{10000} epochs. The FNO model achieves slightly lower training loss, but equivalent validation loss compared to PF-PINO, indicating potential overfitting. Figure~\ref{fig:corrosion1d_ac_ch_loss} presents the one-step training loss curves of the physical residuals for the Allen-Cahn and Cahn-Hilliard equations during PF-PINO training. The Allen-Cahn residual loss consistently decreases during training, while the Cahn-Hilliard residual loss rapidly drops in the early stages and slightly increases later, suggesting a trade-off in minimising the two PDE residuals.



\begin{figure}[htbp]
    \centering
    \includegraphics{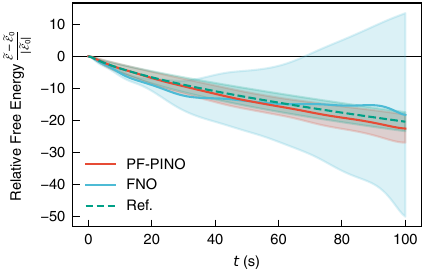}
    \caption{Pencil-electrode corrosion: Normalised
    boundary-work-corrected energy change during 100-step autoregressive prediction.
    Curves show the mean across the six unseen test cases, and shaded bands
    indicate one standard deviation.}
    \label{fig:corrosion1d_corrected_energy}
\end{figure}

To assess thermodynamic consistency
during autoregressive prediction, we evaluate the energy balance associated
with free-energy dissipation.
Because fixed Dirichlet boundary values are prescribed for the concentration
throughout the simulation, the domain can exchange chemical power with its
surroundings.
To separate internal dissipation from this boundary contribution, we subtract
the cumulative boundary chemical work from the free energy in
Equation~\eqref{eq:total_free_energy}. With
$\mu_c=\delta\mathcal{E}/\delta c$, the cumulative boundary work is
\begin{equation}
    W_{\mathrm{bdry}}(t)
    = \int_0^t
    M\left[
    \mu_c(\tau)\partial_x\mu_c(\tau)
    \right]_{x_{\mathrm{L}}}^{x_{\mathrm{R}}}
    \,\mathrm{d}\tau.
\end{equation}
We define the normalised boundary-work-corrected energy change as
\begin{equation}
    \widetilde{\mathcal{E}}_{\mathrm{rel}}(t)
    = \frac{
    \widetilde{\mathcal{E}}(t)-\widetilde{\mathcal{E}}(0)
    }{
    \lvert\widetilde{\mathcal{E}}(0)\rvert
    },
    \qquad \text{where} \qquad
    \widetilde{\mathcal{E}}(t)
    = \mathcal{E}(t)-W_{\mathrm{bdry}}(t).
\end{equation}
The state free energy $\mathcal{E}(t)$ is non-negative. Negative values of
$\widetilde{\mathcal{E}}_{\mathrm{rel}}(t)$ quantify net dissipation after
subtracting the cumulative boundary chemical work. Magnitudes greater than
unity are possible because the boundary chemical work can sustain cumulative
dissipation exceeding the initial stored free energy.
Figure~\ref{fig:corrosion1d_corrected_energy} shows the resulting evolution over the
full autoregressive rollout. PF-PINO follows the non-increasing energy-balance
trend of the FEM reference throughout the rollout. In contrast, the FNO mean
curve exhibits a distinct upward excursion at later time steps, while the upper
edge of its shaded band rises more strongly and crosses zero. These excursions
are inconsistent with the expected non-increasing energy-balance trend.

\subsection{Electro-polishing corrosion}
\label{sec:electro-polishing-corrosion-additional-visualisations}

\begin{figure}[htbp]
    \centering
    \includegraphics{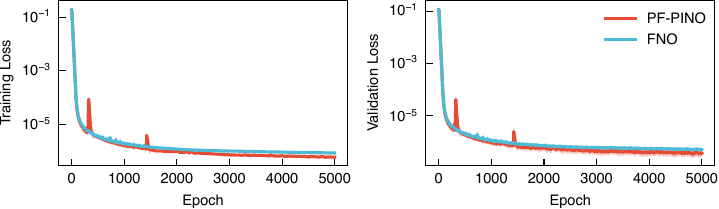}
    \caption{Electro-polishing corrosion: One-step training and validation loss curves during PF-PINO and FNO training. }
    \label{fig:corrosion2d_training_validation_loss}
\end{figure}

\begin{figure}[htbp]
    \centering
    \includegraphics{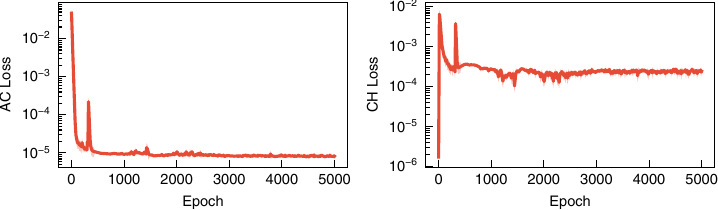}
    \caption{Electro-polishing corrosion: One-step training loss curves of the Allen-Cahn and Cahn-Hilliard equations during PF-PINO training. }
    \label{fig:corrosion2d_ac_ch_loss}
\end{figure}

Figure~\ref{fig:corrosion2d_training_validation_loss} illustrates the one-step training and validation loss curves during PF-PINO and FNO training for the electro-polishing corrosion benchmark. The PF-PINO model achieves slightly lower training and validation loss compared to the FNO. Figure~\ref{fig:corrosion2d_ac_ch_loss} presents the one-step training loss curves of the physical residuals for the Allen-Cahn and Cahn-Hilliard equations during PF-PINO training. Both residual losses rapidly decrease in the initial training stage and gradually stabilise later.

\begin{figure}[htbp]
    \centering
    \includegraphics{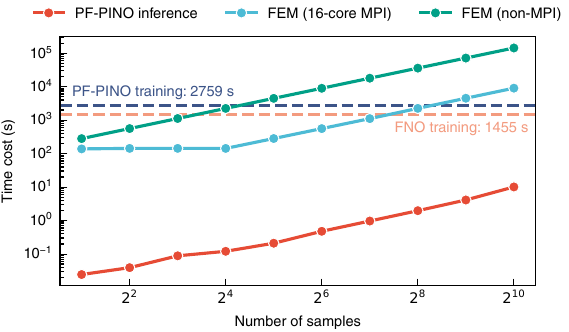}
    \caption{Electro-polishing corrosion: Wall-clock costs of
    PF-PINO and FNO training, PF-PINO inference, and FEM simulation as the
    number of samples increases. PF-PINO inference and FEM simulation times
    correspond to generating complete trajectories at each sample count;
    PF-PINO uses 100 autoregressive updates with
    $\Delta t=\qty{200}{\second}$ per trajectory. Model training and inference
    use one NVIDIA GeForce RTX 5090 GPU, while FEM simulations use either a
    non-MPI process or \num{16}-core MPI parallelisation on a CPU node with an
    Intel Xeon Platinum 8470Q processor.}
    \label{fig:corrosion2d_time_cost}
\end{figure}

To quantify computational cost,
Figure~\ref{fig:corrosion2d_time_cost} reports the wall-clock times for model
training, PF-PINO inference, and FEM simulation for the electro-polishing
corrosion benchmark. For each sample, PF-PINO uses
$\Delta t=\qty{200}{\second}$ to generate the full trajectory from the initial
state to the final time through 100 autoregressive updates. The reported
inference time is the total wall-clock time required to generate the complete
trajectories at each sample count. PF-PINO and FNO training require
\qty{2759}{\second} and \qty{1455}{\second}, respectively. The higher PF-PINO
training cost reflects the additional evaluation of the governing-equation
residuals.

Once trained, PF-PINO inference remains below \qty{10}{\second} for up to
\num{1024} samples. Over the same range, the non-MPI FEM cost increases to
more than \qty{1e5}{\second}, while \num{16}-core MPI parallelisation reduces
this cost to approximately \qty{1e4}{\second}. Thus, PF-PINO inference is
orders of magnitude faster than both FEM implementations across the tested
sample counts. Its one-time training cost can therefore be amortised over
repeated evaluations in parametric studies.

To assess sensitivity to the
autoregressive time step, we independently train PF-PINO models with
$\Delta t$ ranging from \qty{50}{\second} to \qty{400}{\second} in increments
of \qty{50}{\second}. For each time step, the training and validation pairs
are obtained by downsampling the trajectories generated by FEM with
$\Delta t=\qty{50}{\second}$, and the corresponding $\Delta t$ is used in the
PDE residual. All error metrics are evaluated against the FEM trajectories with
$\Delta t=\qty{50}{\second}$ on the same held-out test cases. The FEM
trajectories generated at each tested time step are compared with this common
reference to quantify temporal discretisation error.

\begin{figure}[htbp]
    \centering
    \includegraphics{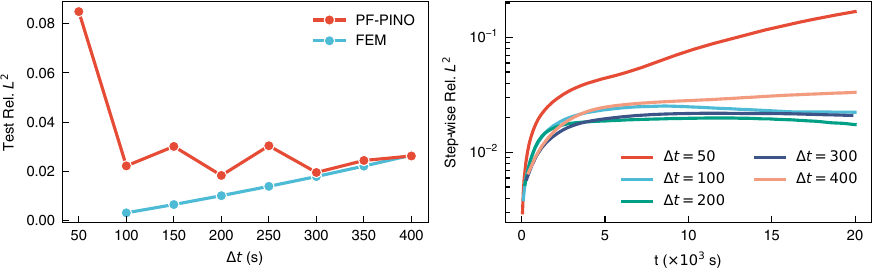}
    \caption{Electro-polishing corrosion: Time-step sensitivity of
    PF-PINO over the full autoregressive prediction horizon.
    The left panel shows the test relative $L^2$ errors of PF-PINO and FEM for
    different time steps, evaluated against FEM trajectories generated with
    $\Delta t=\qty{50}{\second}$. The right panel shows the step-wise test
    relative $L^2$ errors for five representative time steps.}
    \label{fig:corrosion2d_dt_sensitivity}
\end{figure}

The left panel of Figure~\ref{fig:corrosion2d_dt_sensitivity} shows that the
FEM temporal discretisation error increases approximately linearly with
$\Delta t$ and remains lower than \qty{2.6}{\percent} over the tested
range. This indicates that the temporal discretisation error of the backward
Euler scheme remains controlled across the tested time-step range. The
PF-PINO test relative $L^2$ error is approximately
\qty{3}{\percent} or lower for
$\Delta t=\qtyrange{100}{400}{\second}$, corresponding to 50--200
autoregressive steps over the full time horizon. At
$\Delta t=\qty{50}{\second}$, the model requires 400 autoregressive updates,
leading to greater accumulated error, yet the overall test relative $L^2$
error remains below \qty{10}{\percent}. The step-wise errors in the right
panel remain stable after the initial rollout stage for
$\Delta t=\qtyrange{100}{400}{\second}$, demonstrating robust long-horizon
prediction across these time-step choices. The time step used in the main
benchmark, $\Delta t=\qty{200}{\second}$, therefore provides a practical
balance between temporal discretisation accuracy and the number of
autoregressive updates.

\begin{figure}[htbp]
    \centering
    \includegraphics{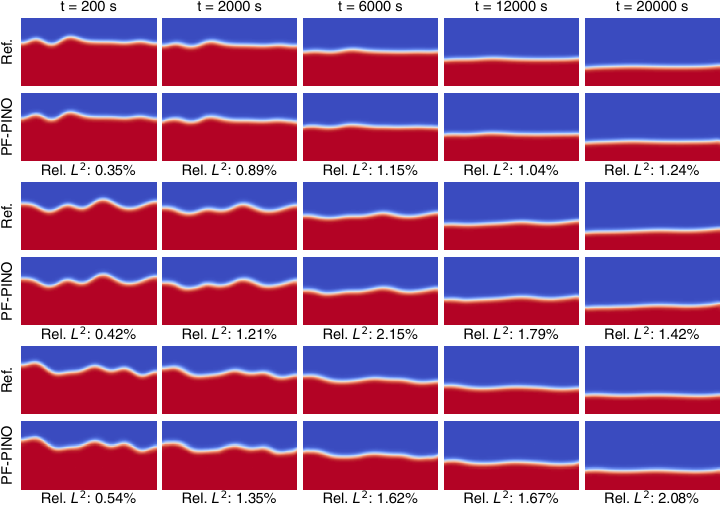}
    \caption{Electro-polishing corrosion: Phase-field variable $\phi$ snapshots of the predicted and ground truth solution fields at different time steps during 100-step autoregressive prediction with different initial interface morphologies. The prediction is obtained via the PF-PINO model.}
    \label{fig:corrosion2d_test_sol_samples}
\end{figure}

We supplement the solution field visualisation with different initial interface morphologies in Figure~\ref{fig:corrosion2d_test_sol_samples}. The predicted phase-field variable $\phi$ snapshots at different time steps during \num{100}-step autoregressive prediction closely match the ground truth simulation results, with a maximum relative $L^2$ error of around \qty{2.0}{\percent} at the final time step. 

\subsection{Dendritic crystal solidification}
\label{sec:dendritic-solidification-additional-visualisations}

\begin{figure}[htbp]
    \centering
    \includegraphics{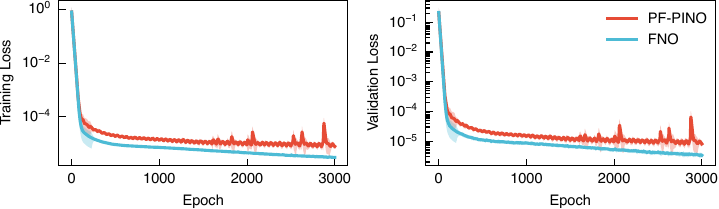}
    \caption{Dendritic crystal solidification: One-step training and validation loss curves during PF-PINO and FNO training. }
    \label{fig:solidification_training_validation_loss}
\end{figure}

\begin{figure}[htbp]
    \centering
    \includegraphics{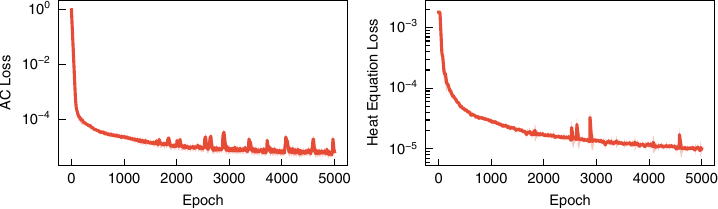}
    \caption{Dendritic crystal solidification: One-step training loss curves of the phase-field Allen-Cahn equation and heat equation during PF-PINO training. }
    \label{fig:solidification_ac_tem_loss}
\end{figure}

Figure~\ref{fig:solidification_training_validation_loss} illustrates the one-step training and validation loss curves during PF-PINO and FNO training for the dendritic crystal solidification benchmark. Both models exhibit stable convergence in \num{3000} epochs. The FNO model achieves slightly lower training and validation loss compared to PF-PINO, but the test results in the main text demonstrate that PF-PINO attains superior long-term prediction accuracy. This result implies that minimising one-step loss does not necessarily guarantee better autoregressive prediction performance, especially for unseen parametric conditions. Figure~\ref{fig:solidification_ac_tem_loss} presents the one-step training loss curves of the physical residuals for the phase-field Allen-Cahn equation and heat equation during PF-PINO training. Both residual losses consistently decrease during training, indicating the effectiveness of the staggered training scheme in mitigating gradient conflicts.


\begin{figure}[htbp]
    \centering
    \includegraphics{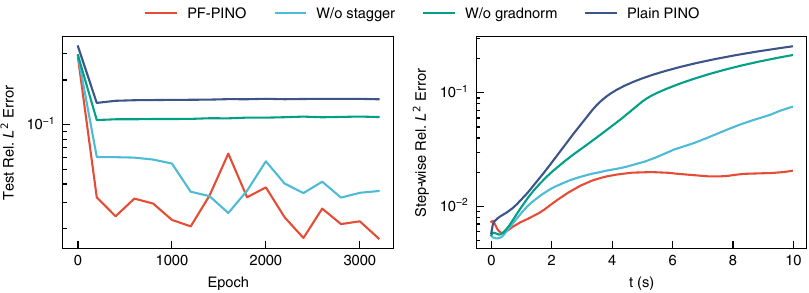}
    \caption{Dendritic crystal solidification: Complete ablation of
    gradient normalisation and staggered training. Left: test relative $L^2$
    error during training. Right: step-wise relative $L^2$ error during
    200-step autoregressive prediction.
    The abbreviations indicate which components are included in each setting:
        PF-PINO uses both gradient normalisation and staggered training;
        W/o stagger uses gradient normalisation but not staggered training;
        W/o GradNorm uses staggered training but not gradient normalisation;
        Plain PINO uses neither gradient normalisation nor staggered training.}
    \label{fig:solidification_test_l2_ablation}
\end{figure}

We expand the ablation study into
a complete $2\times2$ evaluation of gradient normalisation and staggered
training. The four settings are the complete PF-PINO, PF-PINO without staggered
training, PF-PINO without gradient normalisation, and plain PINO without either
strategy. Figure~\ref{fig:solidification_test_l2_ablation} compares the test
relative $L^2$ error during training and the step-wise relative $L^2$ error over
the 200-step autoregressive rollout. Table~\ref{tab:solidification_ablation}
reports the final-epoch test errors.

The complete PF-PINO achieves the lowest final-epoch test error of
\qty{1.72}{\percent}. Removing staggered training increases the error to \qty{2.55}{\percent}, while removing gradient
normalisation increases it to \qty{10.75}{\percent}, which is \num{6.24} times
the complete-model error. Plain PINO reaches \qty{14.02}{\percent}, or
\num{8.14} times the complete-model error. At later rollout times, the four
settings retain the same performance
ordering. PF-PINO maintains the lowest and most stable error, whereas the
variants without gradient normalisation and without either strategy exhibit
pronounced error accumulation. It is worth noting that the plain PINO error is
higher than the \qty{4.85}{\percent} error of the matched FNO baseline in
Table~1, indicating that incorporating PDE residual
losses without addressing gradient imbalance and conflicts can degrade
predictive performance. This ablation highlights the importance of the
accompanying optimisation strategies for accurate and stable long-term
prediction.

\begin{table}[htbp]
    \centering
    \caption{Dendritic crystal solidification: Final-epoch test relative $L^2$
    error for the complete ablation of gradient normalisation and staggered training.
    The abbreviations are the same as in Figure~\ref*{fig:solidification_test_l2_ablation}.
    }
    \label{tab:solidification_ablation}
    \begin{tblr}{
        width=\textwidth,
        colspec={X[1,c]X[1,c]X[1,c]X[1.4,c]},
        row{1}={font=\bfseries}
        }
        \hline
        Setting & Gradient normalisation & Staggered training
        & Test rel. $L^2$ error (in \%) \\
        \hline
        PF-PINO        & \ding{51} & \ding{51} & \num{1.72} \\
        W/o stagger    & \ding{51} & \ding{55} & \num{2.55} \\
        W/o GradNorm   & \ding{55} & \ding{51} & \num{10.75} \\
        Plain PINO     & \ding{55} & \ding{55} & \num{14.02} \\
        \hline
    \end{tblr}
\end{table}

\begin{figure}[htbp]
    \centering
    \includegraphics{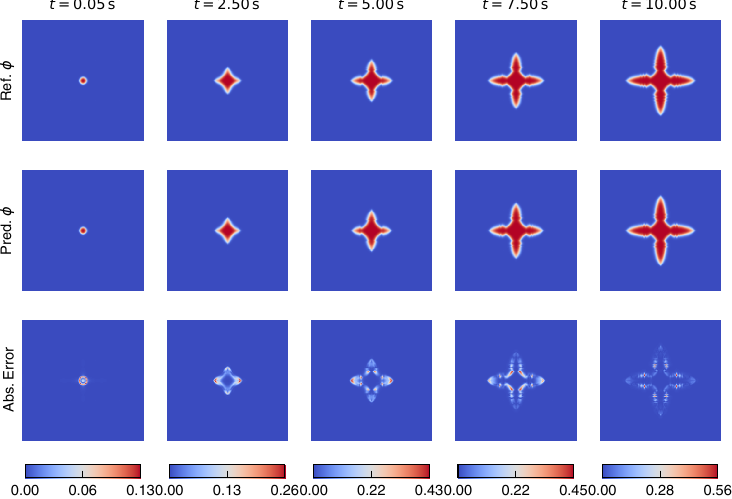}
    \caption{Dendritic crystal solidification: Spatial distribution snapshots of the phase-field variable $\phi$ during solidification at different time steps with the latent heat coefficient $K=1.3$.}
    \label{fig:solidification_test_sol}
\end{figure}

\begin{figure}[htbp]
    \centering
    \includegraphics{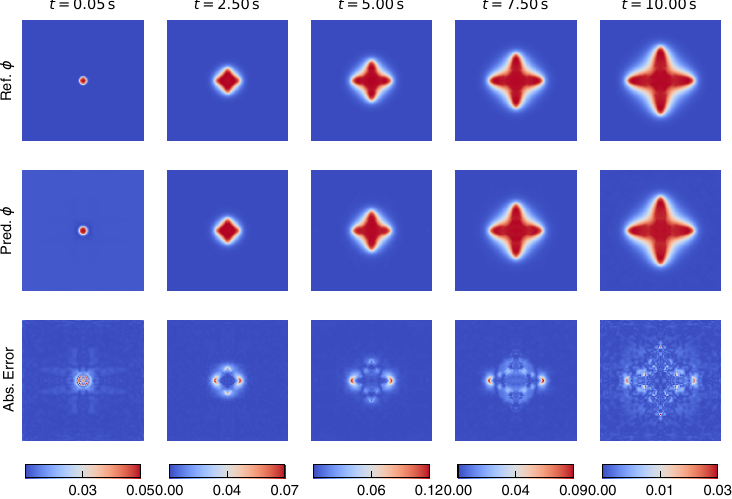}
    \caption{Dendritic crystal solidification: Spatial distribution snapshots of the temperature field $T$ during solidification at different time steps with the latent heat coefficient $K=1.3$.}
    \label{fig:solidification_test_sol_t}
\end{figure}

As a supplementary visualisation to the solution fields at the final rollout time step presented in the main text, Figure~\ref{fig:solidification_test_sol} and Figure~\ref{fig:solidification_test_sol_t} show the spatial distribution snapshots of the phase-field variable $\phi$ and temperature field $T$ at different time steps during solidification with the latent heat coefficient $K=1.3$. The prediction is obtained via \num{200}-step autoregressive prediction using the PF-PINO model. The predicted dendritic growth morphology and temperature evolution closely match the ground truth simulation results. The maximum localised error occurs near the solid-liquid interface due to the steep gradients.

\begin{figure}[htbp]
    \centering
    \includegraphics{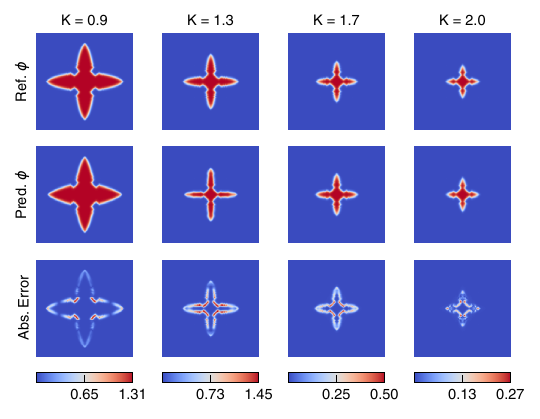}
    \caption{Dendritic crystal solidification: Spatial distribution snapshots of the phase-field variable $\phi$ at the final time step ($t=\qty{10.0}{s}$) predicted by the FNO model trained on a dataset with extended latent heat coefficient range $K \in \{0.8, 1.0, 1.4, 1.6, 2.2\}$.}
    \label{fig:solidification_experiments_expanded_krange_phi}
\end{figure}

To investigate whether expanding the training parameter range can obviate the need for physics-informed constraints, we train a standard FNO on an extended range $K \in \{0.8, 1.0, 1.4, 1.6, 2.2\}$ while maintaining the same number of training samples. As shown in Figure~\ref{fig:solidification_experiments_expanded_krange_phi}, expanding the training range converts the previously extrapolated test cases ($K=1.7,\,2.0$) into interpolation scenarios, which improves their prediction accuracy. However, the increased sampling interval leads to degraded interpolation accuracy for $K=0.9$ and $K=1.3$, exhibiting notable morphological discrepancies compared to the PF-PINO results (Figure~4d in the main text). Maintaining adequate sampling density across a broader parameter range would require proportionally more training data, thereby increasing computational costs. This trade-off underscores the data efficiency of PF-PINO, which achieves robust generalisation without requiring an expanded training dataset.

\subsection{Spinodal decomposition}
\label{sec:spinodal-decomposition-additional-visualisations}

\begin{figure}[htbp]
    \centering
    \includegraphics{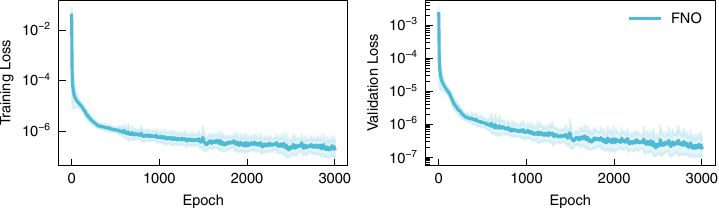}
    \caption{Spinodal decomposition: One-step training and validation loss curves during FNO training.}
    \label{fig:spinodal_decomp_training_validation_loss}
\end{figure}

\begin{figure}[htbp]
    \centering
    \includegraphics{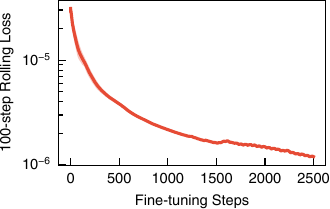}
    \caption{Spinodal decomposition: 100-step autoregressive fine-tuning loss curves during PF-PINO training.}
    \label{fig:spinodal_decomp_fine_tune_loss}
\end{figure}

Figure~\ref{fig:spinodal_decomp_training_validation_loss} illustrates the
one-step training and validation loss curves during FNO training for the
spinodal decomposition benchmark. The FNO model exhibits stable convergence in
\num{3000} epochs. After one-step training, the model undergoes physics-informed
fine-tuning over the entire 100-step autoregressive rollout, during which the
PF-PINO model is trained to minimise the cumulative PDE residual loss.
Figure~\ref{fig:spinodal_decomp_fine_tune_loss} presents the 100-step
autoregressive fine-tuning loss curves during PF-PINO training. The PDE residual
loss consistently decreases during fine-tuning, indicating that the PF-PINO
model effectively learns to satisfy the governing Cahn--Hilliard equation
during long-term prediction.

\begin{figure}[htbp]
    \centering
    \includegraphics{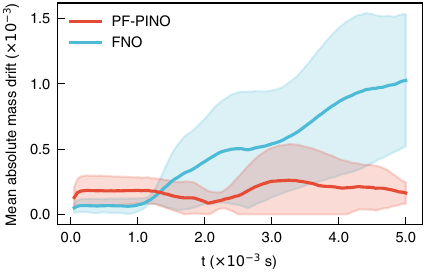}
    \caption{Spinodal decomposition: Absolute concentration-mass
    drift during 100-step autoregressive prediction. The first plotted state
    corresponds to $t=\Delta t$. Curves show the mean across the five unseen
    test cases, and shaded bands indicate one standard deviation.}
    \label{fig:spinodal_decomposition_test_mass_drift}
\end{figure}

We further evaluate
concentration-mass conservation over the autoregressive
rollout.
Under the periodic boundary conditions, the Cahn--Hilliard equation conserves
the total concentration mass. For test case $i$, we quantify the mass drift at
each predicted state $t_n=n\Delta t$, $n=1,\ldots,100$, as
\begin{equation}
    \Delta m_i(t_n)
    =\left\lvert
    \int_\Omega c_i(\bm{x},t_n)\,\mathrm{d}\bm{x}
    -\int_\Omega c_i(\bm{x},0)\,\mathrm{d}\bm{x}
    \right\rvert.
\end{equation}
Figure~\ref{fig:spinodal_decomposition_test_mass_drift} shows the mean and standard deviation
of this quantity across the five unseen test cases. The FNO mass drift grows
progressively during the later rollout and exhibits a clear divergent trend.
At later time steps, the PF-PINO mass drift remains lower and more stable,
without the divergent trend observed for FNO.

\begin{figure}[htbp]
    \centering
    \includegraphics{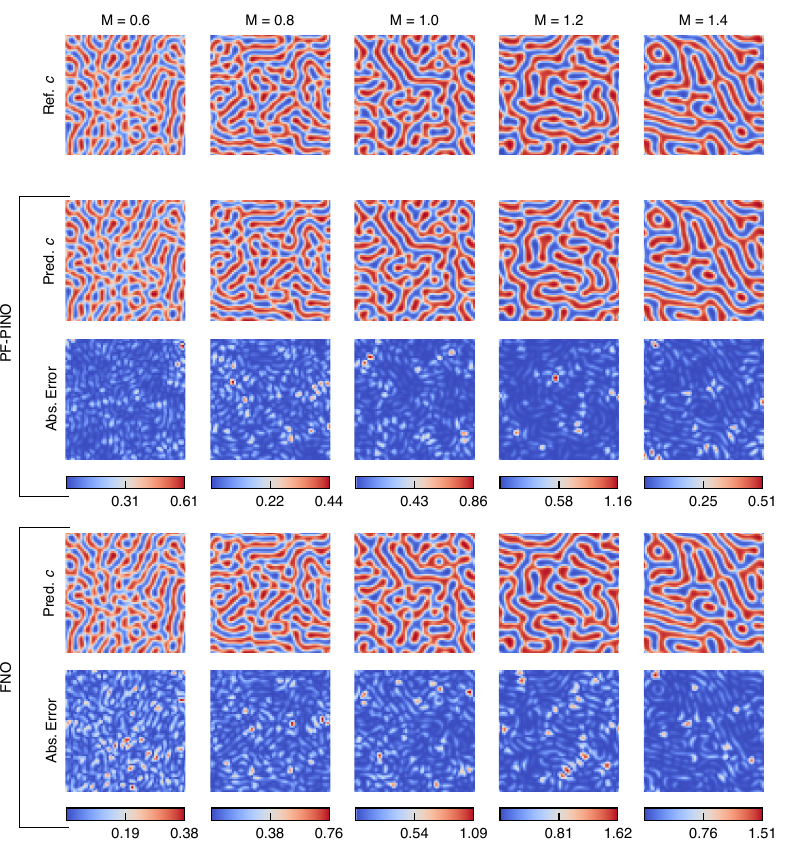}
    \caption{Spinodal decomposition: Spatial distribution of the concentration
    field $c$ at the final time step obtained via 100-step autoregressive
    prediction for different mobilities and initial concentration
    perturbations.}
    \label{fig:spinodal_decomposition_test_final_shape_phi}
\end{figure}

As a supplementary visualisation to the rollout results presented in the main
text, Figure~\ref{fig:spinodal_decomposition_test_final_shape_phi} shows the
spatial distribution of the concentration field $c$ at the final time step
obtained via \num{100}-step autoregressive prediction for the spinodal
decomposition problem with different mobilities and initial concentration
perturbations. The PF-PINO model achieves lower prediction error and better
captures the morphology of the phase separation compared to the FNO model.

\end{document}